%% file: Application_PR_4th.tex
\documentclass[times, review, 10pt]{elsarticle}
\usepackage{amssymb}
\usepackage{amsmath}
 \usepackage{amsthm}

\usepackage{caption}
 \usepackage{subcaption}
 \usepackage{makecell}

 \usepackage{float}
 \usepackage{booktabs}
 \usepackage{array}
 \usepackage{graphicx}%
\usepackage{multirow}%
\usepackage{setspace}
\journal{Pattern Recognition}

\newtheorem{theorem}{Theorem}
\newtheorem{proposition}{Proposition}
\newtheorem{lemma}{Lemma}
\newtheorem{corollary}{Corollary}
\theoremstyle{definition}
\newtheorem{assumption}{Assumption}
\newtheorem{definition}{Definition}
\newtheorem{example}{Example}
\newtheorem{remark}{Remark}

\begin{document}

\begin{frontmatter}
\title{Structure and Regularization of Fused Objectives in Kernel Discriminant Analysis for Small-Sample-Size Problems}

\author[inst1,inst3]{Lingxiao Qu} %% Author name
\ead{d8232113@u-aizu.ac.jp}
%% Author affiliation
\affiliation[inst1]{organization={Graduate School of Computer Science and Engineering, University of Aizu},%Department and Organization
           }
\author[inst2,inst3]{Yan Pei\corref{cor1}}
\ead{peiyan@u-aizu.ac.jp}
\cortext[cor1]{Corresponding author}
\affiliation[inst2]{organization={Computer Science Division, University of Aizu},%Department and Organization
        }
\affiliation[inst3]{addressline={90 Kamiiawase, Tsuruga, Ikki-machi},
            city={Aizuwakamatsu},
            postcode={965-8580},
            state={Fukushima},
            country={Japan}}
%% Abstract
\begin{abstract}
When the number of training samples is smaller than the dimension, discriminant analysis faces a singular within-class scatter matrix, and fused objectives that combine the total variance $C$, the between-class scatter $S_b$ and the within-class scatter $S_w$ are one way to handle it. Using the seven variants of Kernelized Linear Principal Component Discriminant Analysis (KLPCDA) as a platform, we derive the structure of such objectives in this small-sample-size (SSS) setting. The training projections split into a class-mean part and a within-class part, the two scatter operators satisfy $WB=0$, and all between-class information lies in the null space of $S_w$. A single fusion coefficient therefore cannot change the solution, and the variants, KPCA, the regularized kernel Fisher discriminant and the null space methods lie on the boundary of one family with two parameters, the fusion ratio $t$ and a ridge parameter $\mu$. On six tasks from five application domains, the regularized members were more accurate than the best of the seven variants chosen on validation data on all six tasks, by 0.8 to 5.7 points, also under added noise, whereas the fusion ratio gave no gain, including in controlled experiments on dimension ratio, class imbalance and class separation. Selecting a variant per task could gain less than one point, while a fixed choice made on other tasks transferred to every held-out task. The analysis thus gives a principled default for SSS kernel discriminant analysis, with diagnostic quantities, the null space dimension and the class-size spectrum, that can be computed before training.
\end{abstract}

%% Keywords
\end{frontmatter}

%% main text
%%

%% Use \section commands to start a section
\section{Introduction}\label{intro}
The small-sample-size (SSS) problem represents a fundamental challenge across numerous mission-critical machine learning applications where acquiring labeled data is prohibitively expensive, technically constrained or ethically restricted. In hyperspectral remote sensing, for instance, expert annotation of land cover classes requires extensive field surveys under harsh environmental conditions \cite{kumar2024deep}. Similarly, industrial fault diagnosis systems must often operate with limited failure examples due to the high costs of mechanical breakdowns \cite{chen2023deep}, while medical diagnosis is constrained by strict privacy and ethical regulations in collecting large-scale patient data \cite{mahoto2023machine}. Even in face recognition, an extensively investigated computer vision task, many real-world settings such as law enforcement or rare disease screening face the challenge of extremely limited samples per identity \cite{zhang2023application}. These practical constraints render advanced deep learning models, which demand abundant labeled data and complex architectures, often impractical in SSS conditions.

Existing approaches to small-sample learning can be broadly categorized into two paradigms. The first paradigm improves generalization by leveraging knowledge transferred from external datasets, including transfer learning for knowledge reuse~\cite{TLsurvey2023}, few-shot learning for rapid adaptation from few labeled examples~\cite{Tian202407}, meta-learning for learning transferable learning strategies~\cite{Zeng2024679}, and self-supervised pretraining for acquiring general-purpose representations~\cite{Sheshan20252298}. Representative methods include metric-based few-shot learning (e.g., ProtoNet~\cite{snell2017prototypical}), self-supervised representation learning (e.g., SimCLR~\cite{chen2020simple}), and adaptation based on pretrained foundation models~\cite{zhou2025comprehensive}. These approaches have demonstrated strong generalization across a wide range of data-scarce applications, including remote sensing~\cite{TLRS2024}, medical image analysis~\cite{dissanayake2025few}, image recognition~\cite{Zhang20252924}, and agricultural applications~\cite{hossen2025transfer}. However, they generally rely on the availability of large auxiliary datasets and pretrained representations, making them fundamentally different from the classical supervised setting considered in this work. The second paradigm focuses on the classical supervised setting, where only the target training samples are available. Under this setting, dimensionality reduction remains one of the most effective and broadly applicable solutions~\cite{Zhao2024LDA}, particularly when combined with kernel methods to handle linearly inseparable problems~\cite{liu2024face}.

Among the methods developed for the classical supervised setting, Linear Discriminant Analysis (LDA) \cite{LDAIzen} and its variants have been widely used in SSS scenarios to maximize between-class separability while minimizing within-class variances \cite{Qusurvey}. Notable approaches include principal component analysis (PCA) plus LDA, which alleviates the singularity problem through unsupervised dimensionality reduction~\cite{xia2021multiview,PCAplusLDA}; Regularized LDA (RLDA), which improves the stability of covariance estimation by introducing regularization~\cite{RLDA}; Bi-Directional principal component analysis (BDPCA), which preserves two-dimensional structural information for image data~\cite{BDPCAplusLDA}; and KPCA plus LDA, which extends LDA to nonlinear feature spaces through kernel mapping~\cite{KPCAplusLDA}. A second line of work avoids the inverse of the singular within-class scatter matrix by searching the discriminant directions in its null space \cite{Chen2000NLDA,Bodesheim2013KNFST,Ye2006NLDA}, and the regularized kernel Fisher discriminant adds a ridge term to the within-class scatter \cite{Mika1999KFD}. Although these methods address the singularity of the within-class scatter, they are usually used as separate procedures, and how they relate to each other when the objectives are combined is seldom discussed.

Unlike prior-knowledge-driven approaches, kernel discriminant learning does not rely on external pretrained models. Instead, it seeks to improve class separability directly from the limited target samples by constructing discriminative kernel subspaces. Kernelized Linear Principal Component Discriminant Analysis (KLPCDA) \cite{QU2026108539} is a modular discriminant framework of this kind. It fuses the total variance $C$, the between-class scatter $S_b$ and the within-class scatter $S_w$ in a kernel space, and the different ways of combining the three terms give seven variants (Table~\ref{tabKLPCDA}). Compared with a fixed-objective method such as kernel GDA, a fused objective offers more freedom, but it also raises questions that a comparison of accuracies alone cannot answer. First, it is not clear how many of the fusion coefficients can change the learned subspace at all. Second, when the number of training samples $n$ is smaller than the input dimension $m$, $S_w$ is singular, and it is not clear what the fused objectives reduce to in this case, or which of their components is responsible for the accuracy.

The SSS problem is old, and one may ask whether it is still worth studying in the era of big data. We think it is, for a simple reason: what grows with new sensors and new measurement technologies is mostly the number of features, while the number of reliably labeled samples remains limited by cost, expertise and privacy. The gene expression data used in this paper have 49,386 features for 148 samples, and a hyperspectral pixel set with one labeled sample for some classes is not an exception in remote sensing. In this regime the within-class scatter matrix is singular, and the choice of how to handle the singularity decides what the discriminant method actually computes. This is a question about the structure of the objectives, and it is the question of this paper.

In this paper we examine the KLPCDA family from a structural point of view. We show that the three scatter terms of a projection direction depend only on the vector of the training projections, and that this vector splits into a class-mean part and a within-class part. The operators of the between-class and the within-class terms annihilate each other, $WB=0$, so that all the information carried by $S_b$ lies in the null space of $S_w$. From these facts we obtain that a single fusion coefficient cannot change the solution, and that the seven variants are corners and edges of a family with two parameters: the ratio $t$ of the two numerator terms, and the strength $\mu$ of a ridge term on the within-class scatter. We then test these statements on six small-sample tasks and on controlled settings in which the dimension ratio, the class imbalance, the class separation and the noise level are varied one at a time. We evaluate the family on hyperspectral image classification \cite{HyperChau}, fault diagnosis \cite{SMITH2015100}, medical diagnosis \cite{huynh2020improvements}, face recognition \cite{kim2023few} and handwritten digit recognition.

The contributions of this work are threefold:
1) We derive the structure of the fused objectives when the number of samples is smaller than the feature dimension: the decomposition of the training projections, the relation $WB=0$, the null space of $S_w$ and its dimension, and the spectrum of the ratio-type variants, which extends the balanced-class results of \cite{Qu2026Degeneracy} to arbitrary class sizes. Based on these results, a single fusion coefficient is shown not to change the solution, and the seven variants are placed as corners and edges of a family with two parameters. Known methods, namely the null space discriminant methods, the regularized kernel Fisher discriminant and KPCA, belong to the same family.
2) We evaluate the family on six small-sample tasks and in controlled experiments in which one data property is varied at a time. Changing the fusion ratio $t$ between the two edges of the family does not improve the accuracy over the better edge, whereas the ridge parameter $\mu$ does: compared with the best of the seven variants chosen on validation data in every fold, the regularized members achieve a higher mean accuracy on all six tasks, by 0.8 to 5.7 points, with a positive difference for every numerical perturbation seed on five tasks. The advantage remains under added noise, and the regularized members need much smaller subspace dimensions.
3) We show that the variants that apply a pseudo-inverse to the rank-deficient within-class operator are sensitive to numerical perturbations of relative size $10^{-12}$ in the implementation used here, whereas the regularized members are not, and that choosing a variant for each task has a small upper bound and, when done on validation data, costs accuracy. A fixed regularized member is therefore preferable to the selection among variants, and a choice made on other tasks transfers to held-out tasks. Together with the null space dimension and the class-size spectrum, which can be computed before training, this gives a principled default for kernel discriminant analysis in the SSS setting.

The organization of the paper is presented below. Section \ref{meth} introduces the fused objectives and derives their structure in the SSS case. Section \ref{exp} describes the datasets, the compared families and the parameter configuration. Section \ref{ResAnaChapter5} reports the results, the controlled experiments, the numerical sensitivity and the analysis of variant selection. Section \ref{conclu} presents the conclusion and the limitations of this study.

\section{Methodology}\label{meth}
\subsection{Setting and notation}
We consider a training set of $n$ samples $x_i$ with labels in $L$ classes, where class $j$ has $n_j$ samples. A kernel $k$ with feature map $\phi$ gives the kernel matrix $K$, and the centered kernel matrix is $\widetilde{K}=HKH$ with $H=I-\frac{1}{n}\mathbf{1}\mathbf{1}^T$, so that $\widetilde{K}\mathbf{1}=0$ always holds. A projection direction in the centered feature space is written as $v=\sum_{i}\eta_i\tilde\phi(x_i)$ with a coefficient vector $\eta$, and the SSS problem refers to the case $n<m$, where $m$ is the dimension of the input features (the letter $d$ is reserved for the kernel degree). Assuming that the samples are sorted by class, we define
\begin{equation}
D=\mathrm{blkdiag}\Big(\tfrac{1}{n_j}\mathbf{1}\mathbf{1}^T\Big),\quad P=\tfrac{1}{n}\mathbf{1}\mathbf{1}^T,\quad \Lambda=\mathrm{blkdiag}\big(n_jI_{n_j}\big),
\end{equation}
\begin{equation}
B=\tfrac{1}{n}(D-P),\qquad W=\tfrac{1}{n}\Lambda(I-D)=\mathrm{blkdiag}\Big(\tfrac{n_j}{n}I-\tfrac{1}{n}\mathbf{1}\mathbf{1}^T\Big).
\end{equation}
These are the matrices that express the between-class and the within-class scatter in the kernel form of KLPCDA \cite{QU2026108539}. For a direction $v$ the total variance $C$, the between-class scatter $S_b$ and the within-class scatter $S_w$ become
\begin{equation}
v^TCv=\tfrac{1}{n}\eta^T\widetilde{K}^2\eta,\quad v^TS_bv=\eta^T\widetilde{K}B\widetilde{K}\eta,\quad v^TS_wv=\eta^T\widetilde{K}W\widetilde{K}\eta,\quad \|v\|^2=\eta^T\widetilde{K}\eta.
\label{eq:forms}
\end{equation}
The projection of a new sample is computed from its kernel evaluations with the training samples. The classifier in all experiments is the 1-nearest neighbor rule in the projected space.

\subsection{Kernel PCA and kernel discriminant analysis}
KPCA seeks the directions that maximize the variance of the projected data, and its principal components are obtained from the eigenvalue problem
\begin{equation}
\widetilde{K}\eta=n\lambda\eta.
\end{equation}
Details can be found in \cite{KPCA1996}. GDA extends linear discriminant analysis to the kernel space by maximizing $v^TS_bv/v^TS_wv$. With the kernel trick this becomes the generalized eigenvalue problem $\widetilde{K}B\widetilde{K}\eta=\lambda\widetilde{K}W\widetilde{K}\eta$, whose derivation is given in \cite{GDA2000}. The matrix $\widetilde{K}W\widetilde{K}$ on the right-hand side is an $n\times n$ matrix of rank at most $n-L$, so it is singular for every training set, and a classical solution by inversion is not available. How this singularity is handled determines the behavior of the discriminant methods, and we return to this point in Section~\ref{structure}.

\subsection{Fused objectives and the seven variants}
KLPCDA combines three objectives: variance preservation, between-class separability and within-class compactness. Instead of fixing a single objective, it defines a family of models by selecting and combining these terms, with the form
\begin{equation}
\max_v \ \mathcal{J}(v; C, S_b, S_w),
\end{equation}
where $\mathcal{J}$ is a specific combination of the three terms. Table~\ref{tabKLPCDA} lists the seven variants considered in this paper, their target functions, and the operators whose eigenvectors give the projections. In the table, $\alpha$, $\beta$ and $\gamma$ are the fusion coefficients, $t=\beta/\alpha$, and $(\cdot)^{\dagger}$ denotes the Moore--Penrose pseudo-inverse, which replaces the inverse of $W\widetilde{K}$ because this matrix is rank deficient. The last column gives the position of each variant in the two-parameter family that is introduced in Section~\ref{regfamily}.

\begin{table*}[t]
\centering
\renewcommand{\arraystretch}{1.1}
\setlength{\tabcolsep}{3pt}
\caption{Seven variants of KLPCDA: target function, operator whose eigenvectors give the projection, and position in the $(\mu,t)$ family of Section~\ref{regfamily}}
\label{tabKLPCDA}
\resizebox{\textwidth}{!}{
\begin{tabular}{lccll}
\toprule
Method & \makecell{Fused \\Objectives} & \makecell{Target\\ Function} & Operator & Position in the $(\mu,t)$ family \\
\midrule
No.1 & $C$, $S_b$, $S_w$
& $\frac{\alpha v^T C v + \beta v^T S_b v}{\gamma v^T S_w v}$
& $(W\widetilde K)^{\dagger}\big(\frac{1}{n}\widetilde K+tB\widetilde K\big)$
& left edge $\mu=0$, null space block removed$^{*}$ \\

No.2 & $C$, $S_b$
& $\alpha v^T C v + \beta v^T S_b v$
& $\frac{1}{n}\widetilde K+tB\widetilde K$
& edge $\mu\to\infty$, any $t$ \\

No.3 & $S_b$, $S_w$
& $\frac{\beta v^T S_b v}{\gamma v^T S_w v}$
& $(W\widetilde K)^{\dagger}B\widetilde K$
& corner $(\mu=0,\,t\to\infty)$, null space block removed$^{*}$ \\

No.4 & $C$
& $\alpha v^T C v$
& $\frac{1}{n}\widetilde K$
& corner $(\mu\to\infty,\,t=0)$ \\

No.5 & $C$, $S_w$
& $\frac{\alpha v^T C v}{\gamma v^T S_w v}$
& $(W\widetilde K)^{\dagger}\frac{1}{n}\widetilde K$
& corner $(\mu=0,\,t=0)$, null space block removed$^{*}$ \\

No.6 & $S_b$
& $\beta v^T S_b v$
& $B\widetilde K$
& corner $(\mu\to\infty,\,t\to\infty)$ \\

No.7 & $S_w$
& $\gamma v^T S_w v$
& $W\widetilde K$ (smallest eigenvalues)
& invariant to $\mu$; minimal $S_w/\|v\|^2$ \\
\bottomrule
\multicolumn{5}{l}{\footnotesize $^{*}$ With the pseudo-inverse; see Proposition~\ref{prop:spectrum} and Theorem~\ref{prop:family}.}\\
\end{tabular}}
\end{table*}

\paragraph{Method No.1}
This variant considers variance preservation and class discrimination together by combining $C$, $S_b$ and $S_w$. Its objective is
\begin{equation}
\max_v \frac{\alpha v^T C v + \beta v^T S_b v}{\gamma v^T S_w v},
\end{equation}
which leads to the generalized eigenvalue problem $(\alpha C+\beta S_b)v=\lambda\gamma S_wv$. In the kernel representation it reads
\begin{equation}
\big(W\widetilde{K}\big)^{\dagger}\Big(\frac{\alpha}{\gamma n}\widetilde{K}+\frac{\beta}{\gamma}B\widetilde{K}\Big)\eta=\lambda\eta.
\end{equation}
\paragraph{Method No.2}
This variant combines variance and between-class separability without the within-class term, $\max_v \alpha v^T C v + \beta v^T S_b v$, and the solution is the standard eigenvalue problem $(\alpha\frac{1}{n}\widetilde{K}+\beta B\widetilde{K})\eta=\lambda\eta$.
\paragraph{Methods No.3 to No.7}
Methods No.3 and No.5 are the ratio forms $\beta S_b/\gamma S_w$ and $\alpha C/\gamma S_w$, solved as $(W\widetilde{K})^{\dagger}B\widetilde{K}\eta=\lambda\eta$ and $\frac{1}{n}(W\widetilde{K})^{\dagger}\widetilde{K}\eta=\lambda\eta$, respectively. Method No.4 is KPCA and Method No.6 maximizes $v^TS_bv$ under the constraint $\|v\|=1$, which gives the eigenvectors of $B\widetilde{K}$. Method No.7 minimizes the within-class scatter and takes the eigenvectors of $W\widetilde{K}$ with the smallest eigenvalues. Methods No.3 and No.4 are therefore kernel GDA and KPCA written in the notation of the fused objectives.

\subsection{Structure of the problem when $n<m$}\label{structure}
In this subsection we show what the three scatter terms depend on and what the fused objectives reduce to when the within-class operator is singular. The following assumption is used throughout, except where the rank-deficient case is treated explicitly.

\begin{assumption}\label{ass:rank}
The centered kernel matrix has the maximal rank, $\mathrm{rank}\,\widetilde{K}=n-1$.
\end{assumption}

For the polynomial kernel used in this paper, Assumption~\ref{ass:rank} holds when the feature space of the kernel has a dimension of at least $n-1$ and the samples are in general position. Let $U$ be the span of the class indicator vectors, which has dimension $L$ and contains $\mathbf{1}$, and let $U_0=U\cap\mathbf{1}^{\perp}$, which has dimension $L-1$. The matrix $\Pi=D-P$ is the orthogonal projector onto $U_0$, and $I-D$ is the orthogonal projector onto $U^{\perp}$.

\begin{definition}\label{def:split}
For a direction $v=\sum_i\eta_i\tilde\phi(x_i)$, the vector of training projections is $g=\widetilde{K}\eta$, whose entry $g_i$ is the value that $v$ takes on the $i$-th training sample. Its class-mean part and its within-class part are
\begin{equation}
a=\Pi g\in U_0,\qquad b=(I-D)g\in U^{\perp},\qquad g=a+b.
\label{eq:split}
\end{equation}
\end{definition}

Since $\widetilde{K}\mathbf{1}=0$, the vector $g$ always lies in $\mathbf{1}^{\perp}=U_0\oplus U^{\perp}$, and under Assumption~\ref{ass:rank} it can be any vector of this space.

\begin{lemma}\label{lem:forms}
For every direction $v$ with training projections $g=a+b$,
\begin{equation}
v^TCv=\tfrac{1}{n}\big(\|a\|^2+\|b\|^2\big),\qquad v^TS_bv=\tfrac{1}{n}\|a\|^2,\qquad v^TS_wv=\tfrac{1}{n}\,b^T\Lambda b .
\label{eq:abforms}
\end{equation}
\end{lemma}
\begin{proof}
By (\ref{eq:forms}), $v^TCv=\frac{1}{n}g^Tg$, $v^TS_bv=g^TBg$ and $v^TS_wv=g^TWg$. The first two identities follow from $B=\frac1n\Pi$ and the orthogonality of $a$ and $b$. For the third, $g^TWg=\frac1n(a+b)^T\Lambda(I-D)g=\frac1n(a+b)^T\Lambda b$, and $a^T\Lambda b=0$ because $\Lambda a$ is constant on every class while $b$ sums to zero on every class.
\end{proof}

\begin{example}\label{ex:running}
Let $n=6$ and $L=2$, with the first three samples in class 1 and the last three in class 2, so that $n_1=n_2=3$, $\Lambda=3I$, $W=\frac12(I-D)$ and $B=\frac16\Pi$. For $g=(2,4,6,-3,-4,-5)^T$ the class means of the training projections are $4$ and $-4$, hence
\[
a=(4,4,4,-4,-4,-4)^T,\qquad b=(-2,0,2,1,0,-1)^T .
\]
Since $\|a\|^2=96$, $\|b\|^2=10$ and $b^T\Lambda b=30$, Lemma~\ref{lem:forms} gives $v^TCv=106/6$, $v^TS_bv=16$ and $v^TS_wv=5$; the same values follow directly from $Wg=\frac12b$ and $Bg=\frac16a$. The direction with $g=(1,1,1,-1,-1,-1)^T$ has $b=0$, so $v^TS_wv=0$: it maps class 1 to the value $1$ and class 2 to the value $-1$.
\end{example}

\begin{remark}\label{rem:forms}
Lemma~\ref{lem:forms} shows that $S_b$ measures only the class-mean part $a$, $S_w$ only the within-class part $b$, and $C$ both. The three terms are therefore not independent objectives: up to the class-size weights in $W$, the total variance is the sum of the two parts that $S_b$ and $S_w$ measure separately. During training, the data enter only through the norm $\|v\|^2=g^T\widetilde{K}^{\dagger}g$, which couples $a$ and $b$; the projections of new samples are computed from their kernel evaluations as usual. These statements concern any objective built from the three terms when $n<m$, and the seven variants of KLPCDA serve as the experimental platform for them.
\end{remark}

\begin{lemma}\label{prop:wb}
$WB=0$ and $BW=0$. Equivalently, the range of $B$ is contained in the null space of $W$.
\end{lemma}
\begin{proof}
$WB=\frac{1}{n^2}\Lambda(I-D)(D-P)=\frac{1}{n^2}\Lambda\,(D-P-D^2+DP)=0$, because $D^2=D$ and $DP=P$ ($\mathbf 1\in U$). The second identity follows by transposition.
\end{proof}

\begin{remark}\label{rem:balanced}
Lemma~\ref{prop:wb} holds for any class sizes, and it says that all the information carried by the between-class scatter lies in the null space of the within-class operator, $\mathrm{range}(B)=U_0\subset\ker W=U$. Applying Lemma~\ref{prop:wb} with $n_j=k$ for every class, $W=\frac1L(I-D)$ is a multiple of the within-class centering projector, and the lemma reduces to the statement that $B$ vanishes on the range of this projector, which was reported in our preprint \cite{Qu2026Degeneracy}. That is, the balanced case of \cite{Qu2026Degeneracy} is a special case of Lemma~\ref{prop:wb}.
\end{remark}

\begin{lemma}[See \cite{Ye2006NLDA,KPCAplusLDA,Bodesheim2013KNFST}]\label{prop:null}
Under Assumption~\ref{ass:rank}, the nonzero directions with $v^TS_wv=0$ are exactly those with $g\in U_0$. They form a subspace of dimension $L-1$, they map every class to a single point, and inside this subspace $v^TCv=v^TS_bv$. If $\mathrm{rank}\,\widetilde{K}=r<n-1$, the dimension of this subspace is $\max(0,\,r-n+L)$ when $\mathrm{range}\,\widetilde{K}$ and $U_0$ are in general position, and in the coefficients $\eta$ the null space of $W\widetilde{K}$ has dimension $(n-r)+\max(0,r-n+L)$.
\end{lemma}
\begin{proof}
Because $W$ is positive semidefinite, $g^TWg=0$ if and only if $Wg=0$, that is $g\in U$, and with $g\perp\mathbf{1}$ this means $g\in U_0$. Then $b=0$, so $g$ is constant on every class, and $v^TCv=\frac1n\|a\|^2=v^TS_bv$ by Lemma~\ref{lem:forms}. If $\mathrm{rank}\,\widetilde{K}=r$, the admissible $g$ are $\mathrm{range}\,\widetilde{K}\cap U_0$, two subspaces of $\mathbf{1}^{\perp}$ whose dimensions are $r$ and $L-1$ in a space of dimension $n-1$, and generically their intersection has dimension $\max(0,r+L-1-(n-1))$. Finally, $W\widetilde{K}\eta=0$ if and only if $\widetilde{K}\eta\in\mathrm{range}\,\widetilde{K}\cap U_0$, and the vectors $\eta\in\ker\widetilde{K}$, for which $v=0$, contribute the remaining $n-r$ dimensions.
\end{proof}

In Example~\ref{ex:running}, $L-1=1$ and the null space of $S_w$ is spanned by the direction with $g=(1,1,1,-1,-1,-1)^T$. Under Assumption~\ref{ass:rank}, the null space of $W\widetilde{K}$ has dimension $1+(L-1)=L$. Counts of this kind are known for null space discriminant analysis in the linear case \cite{Ye2006NLDA} and in kernel feature spaces \cite{KPCAplusLDA,Bodesheim2013KNFST}; we restate them in the present notation because they give the size of the part of the solution that is shared by all fused objectives, and because the formula of the rank-deficient case is tested in Section~\ref{controlled}.

\begin{remark}\label{prop:scale}
If $E$ is a square matrix and $c\neq0$, then $E$ and $cE$ have the same eigenvectors, and the normalization of the projection vectors removes the remaining scale. Consequently, in the target functions of Table~\ref{tabKLPCDA}, only the ratio $t=\beta/\alpha$ of the two numerator coefficients can change the projections of Methods No.1 and No.2, and for Methods No.3 to No.7, which contain a single coefficient or a single ratio of coefficients, the coefficient only rescales the operator. A search over the fusion coefficients of Methods No.3 to No.7 therefore does not explore different solutions, and any variation of the accuracy over such a search is due to numerical effects in the solution of an ill-conditioned eigenvalue problem. For Methods No.1 and No.2 the search is one-dimensional, and the useful coordinate is $t$.
\end{remark}

\begin{proposition}\label{prop:spectrum}
Suppose that Assumption~\ref{ass:rank} holds. Then the ratio-type variants have the following spectra in exact arithmetic.
\begin{enumerate}
\item[(a)] For Method No.3, the generalized eigenvalue is $+\infty$ with multiplicity $L-1$ (eigenspace $U_0$) and $0$ with multiplicity $n-L$.
\item[(b)] For Method No.5, it is $+\infty$ with multiplicity $L-1$ and, for each class $j$, $1/n_j$ with multiplicity $n_j-1$, with eigenvectors in $U^{\perp}$ supported on class $j$.
\item[(c)] For Method No.1, the eigenvalues on $U_0$ are infinite and those on $U^{\perp}$ are $\alpha$ times the eigenvalues of Method No.5.
\item[(d)] If $(W\widetilde{K})^{\dagger}$ is used in place of the inverse, then $(W\widetilde{K})^{\dagger}B\widetilde{K}=0$, and the operator of Method No.1 equals $\alpha$ times that of Method No.5.
\end{enumerate}
\end{proposition}
\begin{proof}
In the coordinates $g=a+b$ the quadratic forms of Lemma~\ref{lem:forms} are block diagonal with respect to $\mathbf{1}^{\perp}=U_0\oplus U^{\perp}$. On $U_0$, $W$ vanishes while $C=\frac1nI$ and $B=\frac1nI$, so any ratio with $S_w$ in the denominator is infinite. On $U^{\perp}$, $B$ vanishes, $C=\frac1nI$ and $W=\frac1n\Lambda$, which gives the eigenvalues $0$ for $S_b/S_w$ and $1/n_j$ for $C/S_w$, and (c) follows in the same way. For (d), the null space of $(W\widetilde K)^{\dagger}$ is that of $\widetilde KW$, which contains $\ker W=U$ and hence $\mathrm{range}(B\widetilde{K})\subset\mathrm{range}(B)=U_0\subset U$.
\end{proof}

\begin{table}[t]
\centering
\caption{Generalized eigenvalues of the ratio-type variants in exact arithmetic under Assumption~\ref{ass:rank}, inside and outside the null space of $S_w$ (Proposition~\ref{prop:spectrum}), and the effect of a truncated pseudo-inverse.}
\label{tab:spectrum}
\renewcommand{\arraystretch}{1.1}
\setlength{\tabcolsep}{4pt}
\begin{tabular}{lcc}
\toprule
 & \makecell{Inside the null space\\($L-1$ directions)} & \makecell{Outside the null space\\($n-L$ directions)} \\
\midrule
No.3: $S_b/S_w$ & $\infty$ & $0$ \\
No.5: $C/S_w$ & $\infty$ & $1/n_j$, multiplicity $n_j-1$ \\
No.1: $(\alpha C+\beta S_b)/S_w$ & $\infty$ & $\alpha/n_j$, multiplicity $n_j-1$ \\
\midrule
Truncated pseudo-inverse & discarded & kept \\
\bottomrule
\end{tabular}
\end{table}

\begin{example}\label{ex:spectrum}
In Example~\ref{ex:running}, Method No.5 has one infinite eigenvalue, on the direction with $g=(1,1,1,-1,-1,-1)^T$, and the finite eigenvalue $1/3$ with multiplicity $2+2=4$. All finite eigenvalues are equal, so any basis of the four-dimensional space $U^{\perp}$ is an eigenbasis.
\end{example}

\begin{remark}\label{rem:spectrum}
For Indian Pines, whose training class sizes range from 1 to 25, the finite eigenvalues of Method No.5 form a staircase of levels $1/n_j$ (Table~\ref{tab:spectrum}). Applying Proposition~\ref{prop:spectrum} with $n_j=k$ for every class, the finite eigenvalues form a single level, as in Example~\ref{ex:spectrum}; this balanced case, including the vanishing of the operator of Method No.3 and the reduction of Method No.1 to Method No.5 under a truncated pseudo-inverse, was first reported in \cite{Qu2026Degeneracy}, where its consequences for few-shot classification with language-model embeddings are studied. That is, the results of \cite{Qu2026Degeneracy} on the spectra are the balanced special case of Proposition~\ref{prop:spectrum}.
\end{remark}

Outside the null space, the eigenvalues of Methods No.1 and No.5 depend only on the class sizes, and the eigenvectors are the within-class deviations of one class at a time. All the discriminative information of $S_b$ lies in the null space (Lemmas~\ref{prop:wb} and~\ref{prop:null}), so a pseudo-inverse implementation, which discards the infinite eigenvalues, cannot use $S_b$ in Methods No.1 and No.3. In such an implementation the operator $E_3$ of Method No.3 vanishes, and the eigenvectors it returns are determined by rounding errors. On the training sets of Section~\ref{exp}, the ratio $\|E_3\|/\|E_5\|$ of the computed operators of Methods No.3 and No.5 lies between $6\times10^{-15}$ and $6\times10^{-9}$.

\subsection{A regularized family with two parameters}\label{regfamily}
The singularity of $S_w$ can be removed by adding a ridge term to the denominator, as done in the regularized kernel Fisher discriminant \cite{Mika1999KFD}. For the fused objective we replace $v^TS_wv$ by $v^TS_wv+\mu\|v\|^2$ with $\mu>0$. Writing $M=\widetilde{K}^{\dagger}$, which is positive definite on $\mathbf{1}^{\perp}$ under Assumption~\ref{ass:rank}, so that $\|v\|^2=g^TMg$, the objective in the coordinates $g$ is, by Lemma~\ref{lem:forms},
\begin{equation}
\max_{g\in\mathbf{1}^{\perp}}\ \frac{g^TA_tg}{g^T(W+\mu M)g},\qquad A_t=\frac{\alpha}{n}\big(I+t\,\Pi\big),\quad t=\frac{\beta}{\alpha}.
\label{eq:reg}
\end{equation}
For $\mu>0$ the denominator matrix is positive definite, so the stationary points of (\ref{eq:reg}) are the eigenpairs of a symmetric definite pencil. The eigenvalues are real, whereas the nonsymmetric operators of Table~\ref{tabKLPCDA} can return complex eigenvalues and eigenvectors, no pseudo-inverse is needed, and by Remark~\ref{prop:scale} only $t$ and $\mu$ remain as parameters. The main result of this section describes the limits of this family.

\begin{theorem}\label{prop:family}
Suppose that Assumption~\ref{ass:rank} holds. Then:
\begin{enumerate}
\item[(a)] As $\mu\to0^{+}$, $L-1$ eigenvalues diverge as $\lambda_k=\frac{\alpha(1+t)}{n\mu\, m_k}+O(1)$, where $m_k$ are the eigenvalues of the block $M_{aa}$ of $M$ on $U_0$. Their eigenvectors converge to $U_0$, and the basis is determined uniquely by the eigenvectors of $M_{aa}$, ordered by increasing $m_k$. The remaining $n-L$ eigenvalues converge to $\alpha/n_j$ with multiplicity $n_j-1$.
\item[(b)] As $\mu\to\infty$, the solutions converge to those of Method No.2 with ratio $t$; in particular $t=0$ gives KPCA (Method No.4) and $t\to\infty$ gives Method No.6.
\item[(c)] For $t\to\infty$ the numerator is proportional to $\Pi$, so there are exactly $L-1$ nonzero eigenvalues for every $\mu$, and the family member is the regularized kernel Fisher discriminant (RKFD); for $t=0$ it is Method No.5 with a ridge, denoted $\text{No.5}_\mu$.
\end{enumerate}
\end{theorem}
\begin{proof}[Proof sketch]
Write the pencil in the blocks $(a,b)$ of (\ref{eq:split}). $A_t$ is block diagonal with blocks $\frac{\alpha(1+t)}{n}I$ and $\frac{\alpha}{n}I$, $W$ has only the block $W_{bb}=\frac1n\Lambda$ on $U^{\perp}$, and $M$ couples the two blocks. For $\lambda=O(1/\mu)$ the equation of $b$ gives $b=-\mu\,W_{bb}^{-1}M_{ba}a+O(\mu^2)$, and substitution in the equation of $a$ gives the stated expansion. For $\lambda=O(1)$ the equation of $a$ gives $a=O(\mu)$, and the leading order of the equation of $b$ is $\frac{\alpha}{n}b=\lambda\frac{1}{n}\Lambda b$. Part (b) follows from $\mu^{-1}(W+\mu M)\to M$, and part (c) from the rank of $\Pi$. The complete derivation is given in the Supplementary Material.
\end{proof}

In part (a) the limiting directions are the vectors of the null space that maximize $\|a\|^2/(a^TMa)$, that is, the class-mean energy per unit $\|v\|^2$. These are the irregular discriminant vectors of the complete kernel Fisher discriminant framework \cite{KPCAplusLDA}, and they span the same subspace as the null space projections of the kernel null space methods \cite{Bodesheim2013KNFST}. The ridge $\mu\|v\|^2$ is also the natural choice for Method No.5, because with $S_w+\mu C$ the ratio $C/(S_w+\mu C)$ is a monotone function of $C/S_w$ and the eigenvectors do not change. The role of $t$ follows from the same expansion.

\begin{corollary}\label{cor:t}
Suppose that Assumption~\ref{ass:rank} holds. As $\mu\to0^{+}$, the order of the eigenvalues and the limits of the eigenvectors of (\ref{eq:reg}) do not depend on $t$, and the dependence of the eigenvectors on $t$ vanishes at the rate $O(\mu)$. More precisely, the eigenvectors of the diverging block do not depend on $t$ up to terms of order $\mu^2$, and those of the finite block depend on $t$ only through a class-mean component of order $\mu$, scaled by $1/(1+t)$. As $\mu\to\infty$, the eigenvectors converge to those of Method No.2 with ratio $t$, which depend on $t$ at the leading order.
\end{corollary}
\begin{proof}
By the proof of Theorem~\ref{prop:family}, the diverging block has $a$ equal to an eigenvector of $M_{aa}$ and $b=-\mu W_{bb}^{-1}M_{ba}a+O(\mu^2)$, and neither term contains $t$; the factor $1+t$ only scales the diverging eigenvalues, which stay above the finite ones. In the finite block the leading term $b$ is an eigenvector of the restriction of $M$ to a group of classes of equal size, which does not contain $t$, and the equation of $a$ gives $a=\mu\lambda^{(0)}\frac{n}{\alpha(1+t)}M_{ab}b+O(\mu^2)$ (Supplementary Material, S2(b)). The last statement is Theorem~\ref{prop:family}(b).
\end{proof}

\begin{remark}\label{rem:twhere}
Corollary~\ref{cor:t} gives a structural reason why the fusion ratio has little influence for small $\mu$. In the null space $C$ and $S_b$ coincide (Lemma~\ref{prop:null}), so $t$ only rescales the diverging eigenvalues, and outside the null space $S_b$ vanishes (Lemma~\ref{lem:forms}), so $t$ can act only on the small class-mean component that the coupling through $M$ creates. The influence of $t$ grows with $\mu$ and is complete in the limit $\mu\to\infty$. Whether $t$ matters in practice therefore depends on the value of $\mu$ that is selected, which Section~\ref{tratio} examines on the real tasks.
\end{remark}

\begin{remark}\label{rem:multidim}
The criteria above are written for one direction. For $p$ directions we use the eigenvectors of the $p$ largest eigenvalues of the pencil, which solve the ratio-trace problem $\max_V\mathrm{tr}\big((V^T(W+\mu M)V)^{-1}V^TA_tV\big)$, and then normalize each direction to $\|v_k\|=1$; for the seven variants we use the $p$ leading eigenvectors of their operators in the same way. Other extensions of a one-dimensional criterion, such as the trace-ratio criterion $\mathrm{tr}(V^TA_tV)/\mathrm{tr}(V^T(W+\mu M)V)$ or successive orthogonal directions, give different subspaces in general. Lemmas~\ref{lem:forms}--\ref{prop:null} do not depend on this choice, because they concern the scatter forms of each direction and the null space; in particular, any of these criteria with $S_w$ in the denominator is maximized by directions in the null space when $\mu\to0^{+}$. Theorem~\ref{prop:family}, Corollary~\ref{cor:t} and the order of the directions outside the null space are statements about the generalized eigenvalue formulation. The 1-nearest neighbor classifier also depends on the relative scaling of the directions, which is fixed here by the normalization $\|v_k\|=1$.
\end{remark}

\begin{figure}[htbp]
    \centering
    \includegraphics[width=0.78\linewidth]{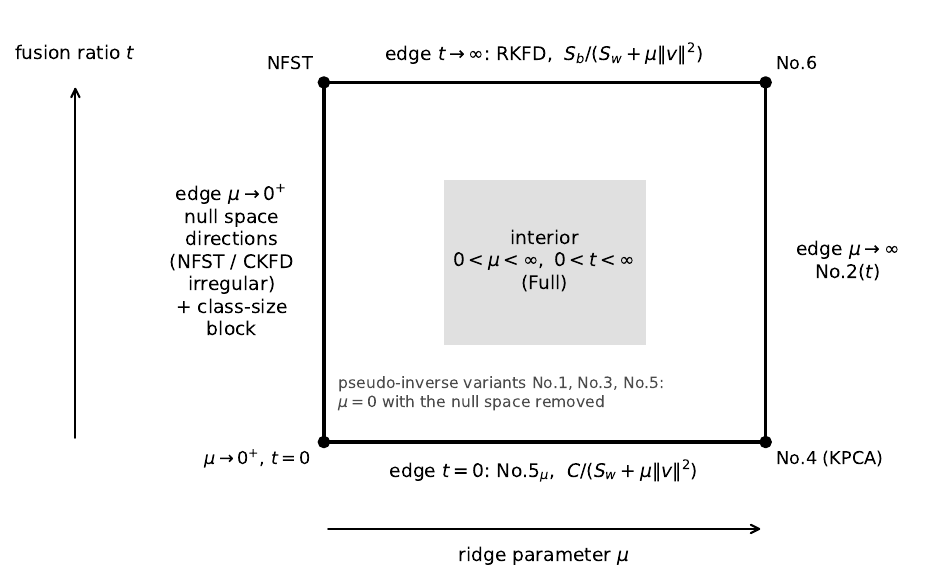}
    \caption{The $(\mu,t)$ family of fused objectives. Rows: fusion ratio $t$; columns: ridge parameter $\mu$. Methods No.4 and No.6 are the corners at $\mu\to\infty$, Method No.2 is the right edge, and the left edge $\mu\to0^{+}$ contains the kernel null space methods. The regularized kernel Fisher discriminant (RKFD) and $\text{No.5}_\mu$ are the top and bottom edges.}
    \label{fig:design}
\end{figure}

Theorem~\ref{prop:family} places the known methods and six of the seven fused variants on the boundary of the square $(\mu,t)\in[0,\infty]^2$ (Table~\ref{tabKLPCDA} and Fig.~\ref{fig:design}). The interior of the square, where both parameters are finite, is the region that the original variants do not reach, and it is explored in Section~\ref{ResAnaChapter5}.

\begin{remark}\label{rem:no7}
Method No.7 is not an element of the square in the strict sense: since $S_w/\|v\|^2$ and $(S_w+\mu\|v\|^2)/\|v\|^2$ are monotonically related, its ordering does not depend on $\mu$. It selects the directions with the smallest within-class scatter relative to $\|v\|^2$. By Lemma~\ref{prop:null}, under Assumption~\ref{ass:rank} the zero eigenspace of $W\widetilde{K}$ has dimension $L$ and consists of the $L-1$ null space directions and the trivial direction $\eta\propto\mathbf{1}$, for which $v=0$. Within this eigenspace the eigenvalue is degenerate, so the basis returned by an eigensolver is arbitrary, and the normalization of the trivial direction divides by $\|v\|=0$. When the subspace dimension $p$ does not exceed $L$, Method No.7 is therefore a null space discriminant with an arbitrary basis, and for larger $p$ it adds the nonzero directions of smallest within-class scatter. For balanced classes, the selection of the null space of $W$ by Method No.7 was also noted in \cite{Qu2026Degeneracy}.
\end{remark}

For linear discriminant analysis on undersampled data, Ye \cite{Ye2005GDA} characterized the maximizers of the criterion $\mathrm{trace}((S_t^L)^{+}S_b^L)$ as a family of transformations $G=X_qM$ with an arbitrary nonsingular $M$, which contains uncorrelated and orthogonal LDA, using a simultaneous diagonalization of $S_b$, $S_w$ and $S_t$. The block structure of Lemma~\ref{lem:forms} plays a similar role in the coordinates of the training projections. The family (\ref{eq:reg}) is different: it is a kernel family, it contains the variance term with a weight, its two parameters are the fusion ratio and the strength of a ridge on $S_w$, and its limits in $\mu$ connect it to the null space methods and to KPCA.

\section{Experiments}
\label{exp}
\subsection{Datasets and Small Sample Setup}
We evaluate the fused objectives on six tasks covering hyperspectral imagery, vibration signals, gene expression data, facial images and handwritten digits. In all tasks the number of training samples is small compared with the feature dimension, which corresponds to the small-sample-size (SSS) setting; for CWRU and MNIST it is close to the number of informative features, so that the centered kernel matrix becomes rank deficient for the linear kernel. Train/validation/test splits are repeated 10 times with fixed random seeds (5 times for MNIST); they are stratified for all tasks except the validation split of MNIST, which is drawn at random. Each task is trained and tested independently, so that the study is a multi-domain evaluation and not a study of cross-domain generalization. Table~\ref{tab:datasets} gives an overview, and the detailed class-wise splits are provided in Tables \ref{IPdatatab}--\ref{mnistdatatab} in the Supplementary Material.

\begin{table}[t]
\centering
\caption{Overview of the six tasks: input dimension $m$, number of classes $L$, number of training samples $n$ and training samples per class.}
\label{tab:datasets}
\renewcommand{\arraystretch}{0.95}
\setlength{\tabcolsep}{3pt}
\resizebox{\columnwidth}{!}{
\begin{tabular}{llcccc}
\toprule
Task & Domain & $m$ & $L$ & $n$ & Per class \\
\midrule
Indian Pines & hyperspectral image & 200 & 16 & 105 & 1--25 (imbalanced) \\
CWRU & bearing vibration & 10 (6 nonconstant) & 4 & 8 & 2 \\
GSE44076 & gene expression & 300 selected of 49,386 & 2 & 15 & 5 and 10 \\
JAFFE-EXP & face images & 4096 & 7 & 42 & 6 \\
JAFFE-ID & face images & at most 59 (in-fold PCA) & 10 & 60 & 6 \\
MNIST & handwritten digits & 784 (594 nonconstant) & 10 & 560 & 51--62 (random split) \\
\bottomrule
\end{tabular}}
\end{table}

\textbf{Indian Pines (Hyperspectral Image Classification).}
The Indian Pines dataset \cite{IP} contains a $145\times145$ hyperspectral image with 224 spectral bands.
After removing water-absorption bands, 200 features remain for 16 imbalanced classes.
A 1\%/5\%/remaining train/validation/test split per class yields 105 training samples versus 200 features, with class-wise training sizes ranging from 1 to 25. Three classes have a single training sample, so their within-class scatter is exactly zero.
Each experiment is repeated 10 times using stratified sampling.

\textbf{CWRU Bearing Dataset (Fault Diagnosis).}
The CWRU bearing dataset \cite{CWRU} provides vibration signals from the drive-end accelerometer under constant operating conditions.
Signals are segmented into windows of 2048 samples, each window is normalized, and 10 time-domain statistical features are extracted from it.
Because the normalization is applied to each window before the statistics are computed, four of the ten features (the mean, the standard deviation, the root mean square and the variance of a normalized window) take the same value in all windows, and the effective feature dimension is six.
Four classes are considered, and 200 samples are split into 8/12/180 training/validation/testing samples.

\textbf{GSE44076 Colon Cancer Dataset (Medical Diagnosis).}
The GSE44076 dataset \cite{GSE44076} provides 49,386 gene expression features, and 148 samples (98 tumor and 50 normal) are used in this paper.
A stratified 10\%/20\%/70\% split yields 15 training samples per run.
Inside every fold, the 300 genes with the smallest $p$-values of the Wilcoxon rank-sum test between the two classes are selected on the training samples only. With 5 and 10 training samples in the two classes the smallest attainable $p$-value is the same for many genes (between 944 and 2081 genes in our folds), so the ties are broken by the absolute difference of the class means divided by the pooled standard deviation. The selected genes are then standardized with the mean and the standard deviation of the training samples, and the same transformation is applied to the validation and test samples.

\textbf{JAFFE Dataset (Face Recognition).}
The JAFFE dataset \cite{jaffe1,jaffe2} contains 213 grayscale facial images from 10 subjects with 7 expressions.
Two tasks are considered: expression and identity recognition.
Images are resized to $64\times64$ and flattened to 4096 gray-level features. For expression recognition these features are used directly. For identity recognition they are reduced by PCA, where the projection basis is computed on the training samples of each fold only and has at most $n-1=59$ components.
For each task, each class is split into 6/5/remaining training/validation/testing samples over 10 repetitions.

\textbf{MNIST (Handwritten Digit Recognition).}
Seven hundred labeled images (70 per class) of the MNIST dataset are used, each with $28\times28=784$ gray-level features, of which 190 are constant on the training samples. In every repetition the 700 images are randomly divided into 560 training and 140 validation images, and a separate set of 700 images is used for testing. The experiment is repeated 5 times.

All experiments are implemented in Python 3 with NumPy and SciPy on a Windows workstation using CPU-only kernel computations. Random splits and random perturbations are controlled by fixed seeds for reproducibility.

\subsection{Compared Families and Evaluation Protocol}
The comparison is made inside the family of fused objectives of Section~\ref{meth}. The first group consists of Methods No.1 to No.7 of Table~\ref{tabKLPCDA}, implemented as described there: the pseudo-inverse of $W\widetilde{K}$ uses the default tolerance $\max(\text{size})\cdot\mathrm{eps}(\|A\|)$, the eigenvalues of the nonsymmetric operators are ordered by modulus and then by phase, the real part of the eigenvectors is kept, each eigenvector is normalized so that $\|v\|=1$, and Method No.7 takes the eigenvalues in ascending order. The second group consists of four members of the regularized family of Section~\ref{regfamily}, all solved as symmetric definite pencils in an orthonormal basis of $\mathrm{range}\,\widetilde{K}$:
\begin{itemize}
\item NFST: the kernel null space limit, realized with $\mu=10^{-4},10^{-3},10^{-2}$ times the smallest scale $\mu_{\mathrm{s}}$ defined below, $t\to\infty$;
\item RKFD: the regularized kernel Fisher discriminant, $t\to\infty$ with $\mu$ selected;
\item $\text{No.5}_\mu$: Method No.5 with the ridge term, $t=0$ with $\mu$ selected;
\item Full: both parameters are selected, $t\in\{0,0.2,1,5,25,\infty\}$.
\end{itemize}
Every family chooses its own configuration, namely the kernel degree $d$, the subspace dimension $p$ and its own parameters, as the one with the highest accuracy on the validation samples of the fold, and the test accuracy of that single configuration is computed once. No test sample is used in any selection. Two summary quantities are used throughout. The best of the seven variants (Best-of-7) is, in every fold, the variant among Methods No.1 to No.7 with the highest validation accuracy, and the best regularized member (Best-of-4) is defined in the same way among the four regularized members. Their difference is the quantity of interest in Section~\ref{ResAnaChapter5}.

\subsection{Parameter Configuration and Optimization}
\subsubsection{Search Ranges and Strategy}
All experiments employ the polynomial kernel
\[
k(x,z)=(\langle x,z\rangle+1)^d,
\]
where $d$ denotes the kernel degree, and the kernel matrix is divided by its largest eigenvalue. Classification performance is evaluated using a 1-nearest neighbor (1-NN) classifier.

The search is performed jointly on the validation samples. The kernel degree $d$ is searched in $\{1,\dots,30\}$ ($\{1,\dots,8\}$ for CWRU and $\{1,\dots,10\}$ for MNIST). The subspace dimension $p$ is searched on a grid of integers from 1 up to the rank of the centered kernel matrix; the grid contains all integers up to 9 and the values 12, 16, 20, 25, 30, 40, 50, 65, 80, 95 and $n-1$ where they do not exceed the rank. For Methods No.1 and No.2, the fusion ratio $t$ is searched on the grid $\{0\}\cup\{10^{-3},10^{-2.4},\dots,10^{3}\}$, because by Remark~\ref{prop:scale} only the ratio is a parameter, and for Methods No.3 to No.7 no coefficient is searched. For the regularized members, $\mu$ is searched on a logarithmic grid of 13 values (11 for MNIST) between $10^{-2}\mu_{\mathrm{s}}$ and $10^{2}\mu_{\mathrm{l}}$, where $\mu_{\mathrm{s}}=(2/n)/\lambda_{\max}(M)$ and $\mu_{\mathrm{l}}=(n_{\max}/n)/\lambda_{\min}(M)$ are defined by the extreme eigenvalues of $M=\widetilde{K}^{\dagger}$ and the largest class size $n_{\max}$, so that the grid covers the range from the null space limit to the $\mu\to\infty$ limit. Table~\ref{pararange} summarizes the search ranges.
\begin{table}[htb]
\centering
\caption{The parameter search ranges across tasks}
\label{pararange}
\setlength{\tabcolsep}{3pt}
\renewcommand{\arraystretch}{0.9}
\begin{tabular}{lccc}
\toprule
Task&\makecell{Polynomial \\degree $d$}  & \makecell{Rank of $\widetilde K$ \\(upper bound of $p$)}&\makecell{Number of\\ $\mu$ values}  \\
\midrule
Indian Pines& $\{1,\dots,30\}$ &104 &13 \\
CWRU & $\{1,\dots,8\}$  & 6--7 &13 \\
GSE44076& $\{1,\dots,30\}$ & 14 & 13\\
JAFFE (identity)& $\{1,\dots,30\}$ &59 & 13\\
JAFFE (expression)& $\{1,\dots,30\}$ &41 & 13\\
MNIST & $\{1,\dots,10\}$ & 552--559 &11 \\
\bottomrule
\end{tabular}
\end{table}

\subsubsection{Numerical Perturbation Protocol}
To measure how reproducible each family is in finite precision, the three feature matrices (training, validation, test) of every fold are multiplied entrywise by $1+10^{-12}\epsilon$ with $\epsilon\sim\mathcal{N}(0,1)$, which is far below any measurement precision, and the whole procedure is repeated with five random seeds. For each family and task we report the mean accuracy over the seeds and the band, defined as the difference between the largest and the smallest mean accuracy over the seeds. All accuracies in Section~\ref{ResAnaChapter5} are means over these five seeds unless stated otherwise.

\section{Results and Analysis}\label{ResAnaChapter5}
All accuracies in this section are test accuracies of the configuration selected on the validation samples, averaged over the folds and over five numerical perturbation seeds (Section~\ref{exp}). The result tables are generated directly from the stored results. For GSE44076 and JAFFE identity recognition, the in-fold preprocessing described in Section~\ref{exp} is used unless stated otherwise.

\input{tables_results}

\subsection{Comparison of the seven variants and the regularized members}\label{main}
Table~\ref{tab:main} reports the accuracy of the eleven families on the six tasks. The comparison of interest is between Best-of-4, the regularized member chosen on validation data in each fold, and Best-of-7, the variant chosen in the same way among the seven variants. Best-of-7 is a deliberately strong opponent, because it can take a different variant in every fold.

On all six tasks the mean accuracy of Best-of-4 is higher than that of Best-of-7. The difference is $+3.22$ points on Indian Pines (10 of 10 folds), $+0.78$ on CWRU (7 of 10), $+1.11$ on GSE44076 (8 of 10), $+5.72$ on JAFFE expression recognition (9 of 10), $+1.69$ on JAFFE identity recognition (7 of 10) and $+1.93$ on MNIST (4 of 5). On five tasks the difference is positive for every perturbation seed. On CWRU the range over the seeds, from $-0.83$ to $+3.89$ points, contains zero, and the comparison on this task is not decided by the present data; Section~\ref{stability} shows that this range is caused by the numerical sensitivity of the variants and not by the regularized members. The size of the difference also depends on the task. GSE44076 is close to saturation, and one point corresponds to about one test sample, whereas JAFFE expression recognition, the most difficult task, shows the largest difference.

Among the seven variants, Method No.7 has the highest accuracy on both JAFFE tasks (40.81\% and 95.90\%), on GSE44076 (98.64\%, shared with the regularized members) and on Indian Pines (55.07\%). By Section~\ref{regfamily}, Method No.7 is a null space discriminant with an arbitrary basis when $p\le L$. The regularized members, whose first $L-1$ directions are the same null space with a uniquely determined basis (Theorem~\ref{prop:family}(a)), are higher still: on JAFFE expression recognition NFST reaches 43.44\% and RKFD 43.91\%, and on JAFFE identity recognition 97.42\% and 97.50\%. The good performance of Method No.7 on the face recognition tasks is therefore consistent with the structure of Section~\ref{structure}: it is the null space that carries the between-class information (Lemma~\ref{prop:wb}), and the result improves further when the basis inside the null space is fixed by the criterion of Theorem~\ref{prop:family}(a) and the remaining directions are regularized.

Method No.3 has the lowest accuracy of the seven variants on five of the six tasks (49.35\% on Indian Pines and 64.62\% on MNIST, for example). By Proposition~\ref{prop:spectrum}, its operator vanishes in the pseudo-inverse implementation, so the low accuracy is explained by the implementation and not by the absence of the variance term in its objective. The regularized kernel Fisher discriminant, which has the same objective $S_b/S_w$ with a ridge term, is the best single family on Indian Pines (58.55\%) and on JAFFE expression recognition (43.91\%).

Table~\ref{tab:p} reports the selected subspace dimension. Methods No.1 and No.5 select the largest admissible dimension in 54\% and 68\% of the selections on Indian Pines, in 50\% on JAFFE identity recognition, and in all selections on MNIST, where $p=559$. The regularized members select a median dimension between 1 and 50 and almost never reach the upper bound, except on CWRU, where the rank of $\widetilde{K}$ is only 6 or 7. This difference follows from Proposition~\ref{prop:spectrum}: in the pseudo-inverse implementation the directions of Methods No.1 and No.5 are the within-class deviations of one class at a time, ordered by the class size only, and many of them are needed before the class means are separated. A regularized member places the null space directions, which separate the class means, first.

Finally, the in-fold preprocessing changes the numbers but not the conclusion. With the PCA basis and the standardization computed on all samples (Table~\ref{tab:noise}, $\sigma=0$), the difference between Best-of-4 and Best-of-7 is $+2.68$ points on JAFFE identity recognition and $+1.07$ points on GSE44076, compared with $+1.69$ and $+1.11$ points with the in-fold preprocessing of Table~\ref{tab:main}.

\subsection{The fusion ratio $t$}\label{tratio}
By Remark~\ref{prop:scale}, the only coefficient that can change the solution of a fused objective is the ratio $t$ of the two numerator terms. Table~\ref{tab:t} compares the member in which both $\mu$ and $t$ are selected (Full) with the better of its two edges, RKFD ($t\to\infty$, numerator $S_b$) and $\text{No.5}_\mu$ ($t=0$, numerator $C$). On no task does Full exceed the better edge on average; the differences lie between $-0.47$ and $0.00$ points. On Indian Pines, Full selects an interior value $0<t<\infty$ in 60\% of the folds, and on MNIST in 28\%, but the accuracy is not higher than at the better edge (by $-0.06$ and $-0.39$ points). Mixing the variance term and the between-class term therefore does not improve the accuracy on the real tasks, even when the validation samples select a mixture.

Corollary~\ref{cor:t} predicts that $t$ can change the solution appreciably only when $\mu$ is not small. To check this, we measured the position of the selected $\mu$ in its range by $s=\log(\mu/\mu_{\mathrm{s}})/\log(\mu_{\mathrm{l}}/\mu_{\mathrm{s}})$, with $s=0$ at the null space scale $\mu_{\mathrm{s}}$ and $s=1$ at the scale $\mu_{\mathrm{l}}$ of the $\mu\to\infty$ limit (Section~\ref{exp}). On GSE44076, CWRU and both JAFFE tasks, the median position selected by Full is between $s=-0.69$ and $s=-0.42$, that is, at the small-$\mu$ end of the range, and Full selects an interior $t$ in at most 8\% of the folds. On Indian Pines and MNIST the selected $\mu$ is larger (median $s=0.44$ and $0.68$), and these are the two tasks on which interior values of $t$ are selected (in 60\% and 28\% of the folds); on MNIST, the folds with an interior $t$ also have a larger $\mu$ (median $s=0.84$) than the folds with $t$ at an edge ($s=0.59$). These observations agree with Corollary~\ref{cor:t}. They also show that a value of $t$ that changes the subspace does not necessarily change the accuracy, since on Indian Pines and MNIST Full is not more accurate than its better edge. Whether an interior $t$ is selected showed no consistent dependence on whether the selected subspace dimension exceeds $L-1$.

Which of the two edges is better depends on the task. RKFD is higher than $\text{No.5}_\mu$ by 1.37 points on Indian Pines and by 0.44 points on JAFFE expression recognition, and lower by 0.66 points on MNIST; on the other three tasks the two are within 0.35 points. The controlled experiments of Section~\ref{controlled} show that this choice is not explained by the class imbalance.

The same consideration applies to the original variants. For Methods No.3 to No.7 a search over the fusion coefficient does not explore different solutions (Remark~\ref{prop:scale}), and for Methods No.1 and No.2 the accuracy must be constant along every line of constant ratio in the plane of the two coefficients. A diagonal structure of the accuracy in that plane, which is observed when both coefficients are swept on a logarithmic grid, is the expected consequence of this invariance, and a variation of the accuracy along a line of constant ratio can only come from the numerical solution of the eigenvalue problem.

\subsection{Controlled experiments}\label{controlled}
The real tasks differ in many properties at once. To isolate the factors, the dimension ratio, the class imbalance, the class separation (overlap) and the noise level were varied one at a time with the other factors fixed. These experiments compare the members of the regularized family; the noise experiment additionally includes all seven variants.

\paragraph{Dimension ratio and rank deficiency.}
Training sets of size $n\in\{105,210,420,840\}$ were drawn from the Indian Pines pool by stratified sampling (the actual sizes are slightly larger because every class receives at least two samples), with the original validation split for model selection and the remaining pool for testing, three repetitions, and the kernel degree fixed at $d=1$ or $d=2$. For $d=1$ the rank of $\widetilde{K}$ is at most 200, so the null space shrinks and finally vanishes as $n$ grows; for $d=2$ the kernel matrix keeps the maximal rank. Table~\ref{tab:dimratio} shows two results. First, the dimension of the null space predicted by Lemma~\ref{prop:null}, $\max(0,r-n+L)$, agrees with the measured dimension in all 24 fold-cells, including the cells where it is zero. On the six real tasks, the dimension of the null space of $W\widetilde{K}$ equals $(n-r)+\max(0,r-n+L)$ in all 165 combinations of fold and kernel degree $d\in\{1,2,3\}$, where the first term counts the directions in $\ker\widetilde{K}$, for which $v=0$. Second, the mixed objective (Full) is not better than the better edge, including the regime where the null space has vanished and all discriminant directions come from the range of $S_w$. In this experiment RKFD is higher than $\text{No.5}_\mu$ in all eight cells, by 0.7 to 4.0 points. Increasing $n$ also makes the task easier, so in this experiment the dimension ratio and the difficulty change together.

\begin{table}[t]
\centering
\caption{Dimension ratio on Indian Pines (three repetitions). $r$: rank of $\widetilde K$; null dimension predicted by Lemma~\ref{prop:null} and measured; test accuracy (\%) of Full, RKFD and $\text{No.5}_\mu$; and Full minus the better edge (points).}
\label{tab:dimratio}
\renewcommand{\arraystretch}{0.9}
\setlength{\tabcolsep}{3pt}
\resizebox{\columnwidth}{!}{
\begin{tabular}{cccccccc}
\toprule
$d$ & $n$ & $r$ & \makecell{Null dim.\\pred./meas.} & Full & RKFD & No.5$_\mu$ & \makecell{Full $-$\\better edge} \\
\midrule
1 & 110 & 109 & 15 / 15 & 58.46 & 59.08 & 58.00 & $-0.61$ \\
1 & 214 & 200 & 2 / 2 & 62.76 & 63.29 & 61.02 & $-0.53$ \\
1 & 424 & 200 & 0 / 0 & 69.95 & 70.11 & 66.87 & $-0.16$ \\
1 & 842 & 200 & 0 / 0 & 74.07 & 74.04 & 71.38 & $+0.03$ \\
2 & 110 & 109 & 15 / 15 & 58.99 & 58.67 & 57.96 & $+0.32$ \\
2 & 214 & 213 & 15 / 15 & 64.67 & 65.21 & 63.37 & $-0.54$ \\
2 & 424 & 423 & 15 / 15 & 72.67 & 73.16 & 69.16 & $-0.49$ \\
2 & 842 & 841 & 15 / 15 & 78.73 & 79.15 & 75.91 & $-0.42$ \\
\bottomrule
\end{tabular}}
\end{table}

\paragraph{Class imbalance and class separation.}
Synthetic data were drawn from Gaussian classes in dimension 200 with $L=8$ classes, 80 training samples, an identity within-class covariance shared by all classes, and 15 random repetitions. The validation and test sets are balanced (10 and 60 samples per class), so the imbalance does not affect the evaluation. On the first axis the imbalance ratio $n_{\max}/n_{\min}$ was set to 1, 2, 4, 8 and 16 with the total size fixed; on the second axis the class separation was set to 0.15, 0.20, 0.25 and 0.35 with balanced classes, which changes the overlap of the classes. The separation of the first axis (0.25) was calibrated so that the accuracy is between the chance level and saturation. Table~\ref{tab:synthetic} reports the results. The difference between the two numerators, RKFD minus $\text{No.5}_\mu$, is negative at every imbalance ratio and shows no trend with the ratio; its largest magnitude occurs for balanced classes. The class imbalance therefore does not decide which numerator is preferable. On the separation axis this difference also shows no trend. Full is below the better edge in 6 of the 9 cells and above it in the other 3, by at most 0.81 points.

Over the 17 controlled cells of Tables~\ref{tab:dimratio} and~\ref{tab:synthetic}, Full is below the better edge in 12 cells (by up to 2.75 points) and above it in 5 cells (by at most 0.81 points). Together with Table~\ref{tab:t}, this shows that the fusion ratio $t$ does not give a systematic gain in any of the regimes examined.

The KPCA corner (Method No.4) reaches only 16.69\% to 74.57\% on the synthetic data, far below the other members, while on the real tasks it is competitive (for example 86.86\% on MNIST). With an isotropic within-class covariance the directions of largest variance carry no class information. On the real tasks the within-class covariance is strongly anisotropic, and the directions of large variance are partly discriminative. This explains why the corner $(\mu\to\infty,t=0)$ can be competitive on real data although it does not use the labels.

\begin{table}[t]
\centering
\caption{Controlled synthetic experiments (15 repetitions): test accuracy (\%) and differences (points). Upper part: imbalance ratio $\rho=n_{\max}/n_{\min}$ (separation 0.25); lower part: class separation (balanced classes).}
\label{tab:synthetic}
\renewcommand{\arraystretch}{0.9}
\setlength{\tabcolsep}{3pt}
\resizebox{\columnwidth}{!}{
\begin{tabular}{ccccccc}
\toprule
Setting & No.4 (KPCA) & RKFD & No.5$_\mu$ & Full & \makecell{RKFD $-$\\No.5$_\mu$} & \makecell{Full $-$\\better edge} \\
\midrule
$\rho=1$ & 36.83 & 66.61 & 69.22 & 66.47 & $-2.61$ & $-2.75$ \\
$\rho=2$ & 36.00 & 67.17 & 67.60 & 66.96 & $-0.43$ & $-0.64$ \\
$\rho=4$ & 36.57 & 58.63 & 59.19 & 60.00 & $-0.57$ & $+0.81$ \\
$\rho=8$ & 30.46 & 47.79 & 48.56 & 48.18 & $-0.76$ & $-0.38$ \\
$\rho=16$ & 32.88 & 42.81 & 42.88 & 43.04 & $-0.07$ & $+0.16$ \\
\midrule
sep. 0.15 & 16.69 & 30.43 & 30.08 & 30.61 & $+0.35$ & $+0.18$ \\
sep. 0.20 & 24.82 & 47.90 & 49.96 & 48.26 & $-2.06$ & $-1.70$ \\
sep. 0.25 & 36.58 & 70.10 & 70.44 & 70.12 & $-0.35$ & $-0.32$ \\
sep. 0.35 & 74.57 & 93.82 & 94.29 & 93.81 & $-0.47$ & $-0.48$ \\
\bottomrule
\end{tabular}}
\end{table}

\paragraph{Noise.}
Gaussian noise $X\leftarrow X+\sigma s\,\mathcal{N}(0,1)$ was added to the training, validation and test features, where $s$ is the mean standard deviation of the training features, so that no information of the validation or test samples is used. Three noise levels $\sigma\in\{0.5,1,2\}$ and three noise draws were used on the five tasks other than MNIST, with the complete protocol of Section~\ref{exp}. Table~\ref{tab:noise} shows that Best-of-4 remains higher than Best-of-7 at every noise level on every task, by between $+0.35$ and $+5.72$ points. The difference tends to become smaller with increasing noise on Indian Pines and on both JAFFE tasks, and it does not change systematically on CWRU and GSE44076. Method No.1 loses more accuracy than RKFD from $\sigma=0$ to $\sigma=2$ on four of the five tasks: 35.1 against 16.5 points on JAFFE identity recognition, 26.8 against 23.0 on CWRU, 19.7 against 17.6 on Indian Pines and 1.6 against 1.3 on GSE44076. On JAFFE expression recognition the order is reversed (14.1 against 19.9 points), although RKFD stays above Method No.1 at every noise level.

\subsection{Numerical sensitivity}\label{stability}
Table~\ref{tab:band} reports the band of every family, that is, the change of its mean accuracy when the input features are perturbed by a relative amount of $10^{-12}$. Methods No.2 and No.4 have a band of exactly zero on all tasks, and Method No.6 has a band below one point. The variants that use the pseudo-inverse of $W\widetilde{K}$ and Method No.7 are sensitive: Method No.1 has a band of 4.50 points on CWRU and 3.53 points on JAFFE expression recognition, Method No.5 a band of 5.56 points on CWRU, and Method No.7 bands up to 2.57 points. On CWRU, the band of Method No.1 is larger than its standard deviation over the folds (2.49 points). The regularized members have bands below 0.5 points on five tasks, with the exception of NFST on CWRU (1.06 points) and MNIST (1.11) and Full on MNIST (0.97); on JAFFE expression recognition their bands are between 1.40 and 2.87 points. We did not investigate the cause of the bands of the regularized members on this task further.

The structure of Section~\ref{structure} explains the sensitive cases. In the pseudo-inverse implementation, the finite eigenvalues of Methods No.1 and No.5 are $\alpha/n_j$ with multiplicity $n_j-1$ (Proposition~\ref{prop:spectrum}); they depend only on the class sizes, so the eigenvalue problem has large degenerate eigenspaces in which any basis is a solution, and a perturbation of the order of the rounding errors can change the returned basis and with it the nearest neighbors. The operator of Method No.3 is zero in exact arithmetic, so its eigenvectors are entirely determined by rounding. Method No.7 works in the zero eigenspace of $W\widetilde{K}$, which is degenerate and contains the trivial direction $\eta\propto\mathbf{1}$. Methods No.2 and No.4 have no such degeneracy, and the regularized members solve a symmetric definite pencil whose eigenvalues are generically distinct. These bands are properties of the implementation described in Section~\ref{exp}; other implementations of the same objectives may behave differently. For the comparison of Section~\ref{main} they mean that a difference between a sensitive variant and another family is meaningful only when it exceeds the band, which is why every number in this section is an average over five perturbation seeds.

\subsection{Choosing a family for each task}\label{selection}
The variants of a fused objective invite the question of which one to use on a given task. Table~\ref{tab:selection} measures what can be gained by answering it. If in every fold the family with the highest test accuracy among all eleven could be chosen, an oracle that is not implementable, the weighted mean accuracy over the 55 folds would be 76.39\%. Fixing the single best family in hindsight gives 75.48\%, and fixing RKFD on every task gives 75.29\%. The ceiling of any per-fold selection rule is therefore 0.91 points above the best fixed family, and 1.10 points above the fixed RKFD.

The standard practice, selecting the family on the validation samples in every fold, reaches 74.39\%, which is 0.90 points below the fixed RKFD; it is lower than the fixed RKFD on five of the six tasks and higher by 0.28 points on MNIST. The same holds within the seven variants. Selecting among them on the validation samples gives 72.79\%, which is 0.75 points below always using Method No.7, and it is below Method No.7 on four of the six tasks (on JAFFE expression recognition, 37.75\% against 40.81\%). The validation sets have between 12 and 513 samples (at most 140 on five tasks), and the differences between the families on them are often smaller than the variation of the validation accuracy itself, so the selection adds variance without adding information.

Small validation sets also make the validation accuracy of a selected configuration an optimistic estimate of its performance, which limits how reliably the kernel degree, $\mu$ and $p$ can be chosen from so few samples. Averaged over the folds and seeds, the validation accuracy of the configuration selected by Best-of-4 exceeds its test accuracy by 0.69 points on GSE44076, 1.90 on JAFFE identity recognition, 2.79 on Indian Pines and 4.45 on MNIST, but by 10.42 points on JAFFE expression recognition and 17.72 points on CWRU, the two tasks with the lowest accuracies and with 35 and 12 validation samples. For Best-of-7 the overestimate is larger on five of the six tasks (for example 14.99 and 21.33 points on the last two tasks). The test accuracies reported in this paper are not affected by this bias, because the test samples are used only once, after all selections; the bias concerns the validation accuracy as an estimate, and it is a further reason to keep the selection small by fixing the family.

A recommendation of a fixed family must also hold on data that were not used to make it. We therefore repeated the choice in a leave-one-task-out manner: for every task, the family with the highest mean accuracy over the other five tasks was chosen and evaluated on the held-out task. RKFD was chosen for five of the six held-out tasks and the member with both parameters (Full) for JAFFE expression recognition. On every held-out task the chosen family was more accurate than Best-of-7 (for example 58.55\% against 55.36\% on Indian Pines and 43.54\% against 37.75\% on JAFFE expression recognition), and its weighted mean over the 55 folds, 75.22\%, is 0.26 points below the best single family chosen in hindsight (75.48\%) and 2.43 points above Best-of-7 (72.79\%).

We therefore do not derive a rule that maps data characteristics to a variant. On these tasks the potential gain of such a rule is below one point, the selection on validation data loses accuracy, and none of the controlled factors of Section~\ref{controlled} determines the choice between the two numerators. A fixed regularized member is the better practice.

\subsection{Computational Complexity}\label{complexity}
The cost of all families is dominated by eigenvalue problems of size $n$, as summarized in Table~\ref{Stepcomp}. The seven variants form the matrices $B\widetilde{K}$ and $W\widetilde{K}$ and solve one eigenvalue problem of size $n$, with $O(n^3)$ operations and $O(n^2)$ storage. A regularized member computes the eigendecomposition of $\widetilde{K}$ once for each kernel degree, which also costs $O(n^3)$, and then solves one symmetric definite pencil of size $r=\mathrm{rank}\,\widetilde{K}\le n-1$ for each value of $\mu$. The cost scales with the number of training samples $n$ and not with the input dimension $m$, so that all families remain inexpensive in the SSS setting.

\begin{table}[t]
\centering
\caption{Complexity and storage of the implementation procedure}
\label{Stepcomp}
\renewcommand{\arraystretch}{0.9}
\setlength{\tabcolsep}{3pt}
\begin{tabular}{lcc}
\toprule
Steps &Complexities&Storage \\ \midrule
kernel matrix $K$ and centering & $O(n^2m)$ & $O(n^2)$\\
products $B\widetilde K$, $W\widetilde K$ & $O(n^2)$ (block structure) & $O(n^2)$\\
eigendecomposition of $\widetilde K$ (regularized members) & $O(n^3)$ & $O(n^2)$\\
eigenvalue problem or symmetric pencil & $O(n^3)$ & $O(n^2)$\\
\midrule
total & $O(n^2m+n^3)$ & $O(n^2)$\\
\bottomrule
\end{tabular}
\end{table}

\section{Discussion}\label{discussion}
\paragraph{What is general and what is specific.}
The results of Section~\ref{structure} do not depend on KLPCDA. They hold for any discriminant objective built from the total variance, the between-class scatter and the within-class scatter of a kernel method when the number of training samples is smaller than the dimension of the feature space: the three terms depend only on the training projections, they act on the class-mean part and the within-class part of these projections, and the between-class operator lives in the null space of the within-class operator. The seven variants of KLPCDA serve as a concrete platform on which these statements can be tested, because they cover the possible combinations of the three terms. The empirical findings, on the other hand, are limited to the six tasks and the controlled settings of this paper, to the polynomial kernel and to the 1-nearest neighbor classifier.

\paragraph{Relation to known methods.}
The end points of the $(\mu,t)$ family are known methods: KPCA \cite{KPCA1996}, the regularized kernel Fisher discriminant \cite{Mika1999KFD}, the null space discriminant methods \cite{Chen2000NLDA,Ye2006NLDA,Bodesheim2013KNFST} and the irregular discriminant vectors of the complete kernel Fisher discriminant \cite{KPCAplusLDA}. The dimension of the null space in the full rank case also agrees with the known counts \cite{Ye2006NLDA}. What this paper adds is the organization of these methods, together with the fused variants, as members of one two-parameter family; the observation that one of the two parameters, the fusion ratio, does not improve the accuracy in any of the settings examined, while the other, the ridge parameter, does; the explanation of the behavior of the pseudo-inverse variants by the spectrum of Proposition~\ref{prop:spectrum}; and the measurement of what the selection among the members can gain. For balanced classes, part of the structure of Section~\ref{structure}, namely the relation between $B$ and the null space of $W$, the single-level spectrum of the pseudo-inverse variants and the vanishing operator of Method No.3, appeared in our preprint \cite{Qu2026Degeneracy}, which studies few-shot classification on frozen language-model embeddings; the present paper derives the general case, the dimension of the null space and the two-parameter family, and studies the classical SSS setting. A family characterization for linear discriminant analysis on undersampled data was given in \cite{Ye2005GDA}; it is parameterized by a nonsingular transformation and does not contain the variance term or a ridge, so the two families answer different questions. The contribution of this paper is therefore an analysis of a design space, and its practical outcome is a principled default for kernel discriminant analysis with $n<m$, together with quantities that can be computed before training and that predict how a member of the family will behave.

\paragraph{Practical consequences.}
Three quantities can be computed from the training data alone, before any model is fitted. The rank $r$ of $\widetilde{K}$ gives the dimension $\max(0,r-n+L)$ of the null space of $S_w$, that is, the number of discriminant directions on which every class is collapsed to one point (Lemma~\ref{prop:null}); when it is zero, as in the rank-deficient cells of Table~\ref{tab:dimratio}, every discriminant direction must come from the range of $S_w$. The class sizes give the finite spectrum $\alpha/n_j$ of the pseudo-inverse variants (Proposition~\ref{prop:spectrum}), so the size of their degenerate eigenspaces, and therefore their numerical sensitivity, is known in advance. Finally, the extreme eigenvalues of $M=\widetilde{K}^{\dagger}$ fix the range of the ridge parameter between the null space limit and the $\mu\to\infty$ limit, which is the range searched in this paper. With these quantities, a regularized member such as RKFD can be used as a fixed default, with only the kernel parameter, the subspace dimension and $\mu$ chosen on validation data.

\paragraph{On the value of the variance term.}
In the null space the total variance coincides with the between-class scatter (Lemma~\ref{prop:null}), and outside the null space, in the pseudo-inverse implementation, it is the only numerator term that remains (Proposition~\ref{prop:spectrum}). Together with the results on $t$, this means that the variance term does not act as a separate stabilizing objective. Its influence is through the choice of the numerator at the edge $t=0$, which is competitive with the edge $t\to\infty$ on some tasks and not on others. Where KPCA itself is competitive, as on the real tasks of this paper, the reason is the anisotropy of the within-class covariance, as the synthetic experiments show.

\paragraph{Limitations.}
Several limitations should be stated explicitly. (1) The CWRU features are strongly degenerate: four of the ten features are constant, the pool of 200 samples contains only 159 distinct feature vectors, and 32 distinct test vectors occur with more than one label in a fold, so that no deterministic classifier can exceed 81.61\% on average. On this task the comparison between the regularized members and the variants is not decided by the present data. (2) The dimension ratio experiment changes the training set size and with it the difficulty of the task, so the two factors are not separated. (3) The class imbalance and separation experiments use isotropic Gaussian classes; anisotropic, heteroscedastic or heavy-tailed classes were not examined. (4) The noise experiment uses the pooled preprocessing for GSE44076 and JAFFE identity recognition and does not include MNIST. (5) Only the polynomial kernel and the 1-nearest neighbor classifier are used, and MNIST has five folds. (6) The bands of the regularized members on JAFFE expression recognition (up to 2.87 points) were not explained. (7) The numerical sensitivity is a property of the implementation described in Section~\ref{exp}, and other implementations of the same objectives may behave differently.

\section{Conclusion}\label{conclu}
This paper studied fused discriminant objectives that combine the total variance, the between-class scatter and the within-class scatter in a kernel space, with the seven variants of KLPCDA as the experimental platform. When the number of training samples is smaller than the dimension, the training projections split into a class-mean part and a within-class part, the between-class and within-class operators satisfy $WB=0$, and all between-class information lies in the null space of the within-class scatter. From these facts a single fusion coefficient cannot change the solution, the variants that use a pseudo-inverse discard the null space and keep only directions ordered by the class sizes, and the seven variants together with KPCA, the regularized kernel Fisher discriminant and the null space methods lie on the boundary of a family with two parameters, the fusion ratio $t$ and the ridge parameter $\mu$.

On six small-sample tasks the regularized members gave a higher mean accuracy than the best of the seven variants chosen on validation data, on every task, and the difference was positive for every perturbation seed on five of them. The advantage remained under added noise and with in-fold preprocessing, and the regularized members needed much smaller subspace dimensions. The fusion ratio gave no gain over the better edge of the family, on the real tasks or in the controlled experiments in which the dimension ratio, the class imbalance and the class separation were varied. The pseudo-inverse variants and Method No.7 were sensitive to input perturbations of relative size $10^{-12}$ in the implementation used here, and their reported accuracies should be read together with this sensitivity. Choosing a family for each task could gain at most about one point and, when done on validation data, lost accuracy, whereas a fixed choice made on other tasks transferred to every held-out task. For kernel discriminant analysis in the SSS setting, the structure therefore leads to a principled default, a fixed regularized member such as the regularized kernel Fisher discriminant with $\mu$ searched over a range fixed by the spectrum of $\widetilde{K}$, and to diagnostic quantities, the null space dimension and the class-size spectrum, that can be computed before training.

Future work will address the choice of the ridge parameter. The structure of Section~\ref{structure} relates the conditioning of the regularized pencil to $\mu$, to the spectrum of $\widetilde{K}$ and to the class sizes, and bounds of this kind could replace the search over $\mu$ by a principled choice. A second direction is the robustness of the class statistics in $B$ and $W$, which are built from sample averages and are therefore sensitive to contaminated or mislabeled samples.

\section*{Acknowledgements}
The authors thank Prof. Takafumi Kanamori (Institute of Science Tokyo) for discussions on hyperparameter selection with small samples, multi-dimensional projections and the role of the fusion parameter.

\bibliographystyle{elsarticle-num}
\bibliography{ApplicationRefer}

\clearpage
\setcounter{table}{0}
\setcounter{figure}{0}
\setcounter{equation}{0}
\section*{Supplementary Material}
This supplementary material provides the class-wise sample distributions of the six tasks, the complete derivation of Theorem~1 and Corollary~1 of the main text, and the results with the pooled preprocessing of GSE44076 and JAFFE identity recognition.
\setcounter{table}{0}
\renewcommand{\thetable}{S\arabic{table}}
\renewcommand{\thefigure}{S\arabic{figure}}
\renewcommand{\theequation}{S\arabic{equation}}

\section*{S1. Data splits}

\begin{table}[htbp]
 \centering
 \caption{Ground truth classes and respective sample number distribution across training, validation, and test splits on the Indian Pines dataset}\label{IPdatatab}
 \setlength{\tabcolsep}{3pt}
  \begin{tabular}{cccccc}
    \toprule
    Class No. & Class & \makecell{Labeled\\Samples}& \makecell{Training\\(1\%)}  & \makecell{Validation\\(5\%)}& \makecell{Testing\\(94\%)} \\
    \midrule
    1 & Alfalfa & 46 &1&2&43 \\
    2& Corn-notill	&1428&14&	71&	1343\\
    3&	Corn-mintill&	830&8	&42	&780\\
    4	&Corn	&237&2	&12&	223\\
    5&	Grass-pasture&	483&5&	24	&454\\
    6&Grass-trees&730&7	&37&	686\\
    7&Grass-pasture-mowed&	28&1	&1&	26\\
    8&	Hay-windrowed	&478&5&	24&	449\\
    9&	Oats&	20&1&	1&	18\\
    10&	Soybean-notill&	972&10&	49	&913\\
    11&	Soybean-mintill	&2455&25&	123&	2307\\
    12&	Soybean-clean	&593&6&	30&	557\\
    13&	Wheat	&205&2&	10&	193\\
    14&	Woods	&1265&13&	63&	1189\\
    15&	Buildings-Grass-Trees-Drives&	386&4&	19&	363\\
    16	&Stone-Steel-Towers&	93&1	&5	&87\\\hline
    &Total& 10249&105&513&9631\\
    \bottomrule
  \end{tabular}
\end{table}

\begin{table}[htbp]
  \centering
   \caption{Classes of 12k DE bearing fault data of CWRU with training, validation and testing samples division}\label{cwrudatatab}
   \setlength{\tabcolsep}{3pt}
  \begin{tabular}{@{}lcccc@{}}
    \toprule
    Class& Segment Pool (Fixed) &Training  & Validation& Testing \\
    \midrule
    Normal&50&2	&	3&45\\
   Inner Race&50&2	&	3&45\\
   Ball&50&2&	3&45\\
   Outer Race&50&2	&	3&45\\ \hline
   Total&200&	8&	12&180\\
    \bottomrule
  \end{tabular}
\end{table}

\begin{table}[htbp]
  \centering
  \caption{Classes of GSE44076 colon cancer data with training, validation and testing samples division}\label{colondatatab}
\setlength{\tabcolsep}{3pt}
  \begin{tabular}{lcccc}
    \toprule
    Class&Sample Size& Training(10\%)  & Validation(20\%)& Testing(70\%) \\
    \midrule
    Normal&50&5	&	10&35\\
  Tumor&98&10	&20	&68\\
\hline
Total&148&	15&30&103\\
    \bottomrule
  \end{tabular}
\end{table}

\begin{table}[htbp]
  \centering
   \caption{Two tasks on JAFFE data with training, validation and testing samples division.}\label{jaffedatatab}
   \setlength{\tabcolsep}{3pt}
  \begin{tabular}{lccccc}
    \toprule    Task&\makecell{Class\\number}&\makecell{Training\\per class} &\makecell{Validation\\per class}&\makecell{Testing\\per class} \\
  \midrule
 Expression Recognition & 7&6&5&all remain\\
  Total&&42&35	&136\\\hline
Identity Recognition& 10&6&5&all remain\\
 Total&&60&50	&103\\
\bottomrule
  \end{tabular}
\end{table}

\begin{table}[htbp]
  \centering
   \caption{MNIST subset with training, validation and testing samples division (five repetitions). The 700 labeled training images (70 per digit) are divided at random into 560 training and 140 validation images in every repetition, so the per-class numbers of these two splits vary (for example, 51 to 62 training images per class in the first repetition).}\label{mnistdatatab}
   \setlength{\tabcolsep}{3pt}
  \begin{tabular}{lcccc}
    \toprule
    Class & \makecell{Labeled\\training pool} & Training & Validation & Testing \\
    \midrule
    each digit 0--9 & 70 & random & random & 70\\\hline
    Total & 700 & 560 & 140 & 700\\
    \bottomrule
  \end{tabular}
\end{table}

\clearpage
\section*{S2. Derivation of Theorem 1 and Corollary 1}
We use the notation of Section~2.4 of the main text. Under $\mathrm{rank}\,\widetilde K=n-1$, every training projection $g\in\mathbf 1^{\perp}$ is written as $g=a+b$ with $a\in U_0$ (dimension $L-1$) and $b\in U^{\perp}$ (dimension $n-L$), and $M=\widetilde K^{\dagger}$ is positive definite on $\mathbf 1^{\perp}$. In the blocks $(a,b)$ the matrices of the pencil $A_tg=\lambda(W+\mu M)g$ are
\begin{equation}
A_t=\begin{pmatrix}\frac{\alpha(1+t)}{n}I&0\\0&\frac{\alpha}{n}I\end{pmatrix},\qquad
W=\begin{pmatrix}0&0\\0&W_{bb}\end{pmatrix},\qquad
M=\begin{pmatrix}M_{aa}&M_{ab}\\M_{ba}&M_{bb}\end{pmatrix},
\end{equation}
where $W_{bb}=\frac1n\Lambda$ restricted to $U^{\perp}$ is positive definite, because $b^T\Lambda b=\sum_jn_j\|b_{(j)}\|^2$. The pencil reads
\begin{align}
\tfrac{\alpha(1+t)}{n}\,a&=\lambda\mu\,(M_{aa}a+M_{ab}b),\label{eq:sa}\\
\tfrac{\alpha}{n}\,b&=\lambda\,(W_{bb}b+\mu M_{ba}a+\mu M_{bb}b).\label{eq:sb}
\end{align}

\paragraph{(a) Divergent eigenvalues as $\mu\to0^{+}$.}
Let $\lambda=\kappa/\mu$ with $\kappa=O(1)$. Multiplying (\ref{eq:sb}) by $\mu$ gives $\mu\frac{\alpha}{n}b=\kappa W_{bb}b+\kappa\mu(M_{ba}a+M_{bb}b)$. At order one, $\kappa W_{bb}b=0$, so $b=O(\mu)$. Writing $b=\mu b_1+O(\mu^2)$, the terms of order $\mu$ give $0=\kappa W_{bb}b_1+\kappa M_{ba}a$, that is
\begin{equation}
b=-\mu\,W_{bb}^{-1}M_{ba}a+O(\mu^2).
\end{equation}
Substituting in (\ref{eq:sa}), the leading order is $\frac{\alpha(1+t)}{n}a=\kappa M_{aa}a$. Hence $a$ is an eigenvector of $M_{aa}$, $M_{aa}a_k=m_ka_k$ with $m_k>0$, and
\begin{equation}
\lambda_k=\frac{\alpha(1+t)}{n\,\mu\,m_k}+O(1).
\end{equation}
The eigenvectors converge to $U_0$, the null space of $S_w$, and the largest eigenvalues belong to the smallest $m_k$. Because $\|v\|^2=a^TM_{aa}a$ and $v^TS_bv=\frac1n\|a\|^2$ for $g=a$, these directions maximize $v^TS_bv/\|v\|^2$ inside the null space of $S_w$, which is the criterion of the irregular discriminant vectors of the complete kernel Fisher discriminant framework. The ratio $t$ only scales these eigenvalues and does not change their eigenvectors or their order.

\paragraph{(b) Finite eigenvalues as $\mu\to0^{+}$.}
Let $\lambda=O(1)$. Equation (\ref{eq:sa}) gives $a=O(\mu)$, so the term $\mu M_{ba}a$ in (\ref{eq:sb}) is $O(\mu^2)$, and the leading order of (\ref{eq:sb}) is $\frac{\alpha}{n}b=\lambda\frac1n\Lambda b$. Let $V_j\subset U^{\perp}$ be the vectors that are supported on class $j$ and sum to zero on it; $\dim V_j=n_j-1$ and $\Lambda=n_jI$ on $V_j$. Therefore
\begin{equation}
\lambda^{(0)}=\alpha/n_j,\qquad\text{with multiplicity } n_j-1\text{ and eigenvectors in }V_j .
\end{equation}
These eigenvalues depend only on the class sizes, and on the corresponding directions the class means of the training projections are equal ($a\approx0$). For the first-order term, let $S_\nu=\bigoplus_{n_j=\nu}V_j$ collect the classes of equal size $\nu$, write $\lambda=\alpha/\nu+\mu\lambda_1+O(\mu^2)$ and $b=b_0+\mu b_1+O(\mu^2)$ with $b_0\in S_\nu$, and take the terms of order $\mu$ in (\ref{eq:sb}): $\frac{\alpha}{n}b_1=\frac{\alpha}{\nu n}\Lambda b_1+\frac{\lambda_1}{n}\Lambda b_0+\frac{\alpha}{\nu}M_{bb}b_0$. Projecting onto $S_\nu$, where the first two terms cancel, gives $0=\frac{\lambda_1\nu}{n}b_0+\frac{\alpha}{\nu}P_\nu M_{bb}b_0$, with $P_\nu$ the orthogonal projector onto $S_\nu$. Hence
\begin{equation}
\lambda=\frac{\alpha}{\nu}-\mu\,\frac{\alpha n}{\nu^2}\,m+O(\mu^2),\qquad m\in\mathrm{eig}\big(P_\nu M_{bb}P_\nu|_{S_\nu}\big),
\end{equation}
so the order inside a group of equal class sizes is determined by the within-class directions with the largest variance per unit $\|v\|^2$. The ratio $t$ enters the finite block only through $a$. From (\ref{eq:sa}), $\frac{\alpha(1+t)}{n}a=\lambda\mu(M_{aa}a+M_{ab}b)$ with $a=O(\mu)$, so that
\begin{equation}
a=\mu\,\lambda^{(0)}\,\frac{n}{\alpha(1+t)}\,M_{ab}b_0+O(\mu^2).
\end{equation}
The eigenvalues of the finite block therefore do not depend on $t$ up to the first order, and its eigenvectors depend on $t$ only through this class-mean component of order $\mu$, scaled by $1/(1+t)$. In the diverging block of part (a), $a$ is an eigenvector of $M_{aa}$ and $b=-\mu W_{bb}^{-1}M_{ba}a+O(\mu^2)$, so its eigenvectors do not depend on $t$ up to terms of order $\mu^2$; this proves Corollary~1.

\paragraph{(c) The limit $\mu\to\infty$.}
Dividing the pencil by $\mu$ and letting $\mu\to\infty$ gives $A_tg=\lambda'Mg$ with $\lambda'=\lambda\mu$. This is the stationarity condition of $g^TA_tg/g^TMg=(\alpha v^TCv+\beta v^TS_bv)/\|v\|^2$, the objective of Method No.2 with ratio $t$; $t=0$ gives KPCA (Method No.4) and $t\to\infty$ gives $\max v^TS_bv/\|v\|^2$ (Method No.6).

\paragraph{(d) The edges $t\to\infty$ and $t=0$.}
For $t\to\infty$, $A_t/t\to\frac{\alpha}{n}\Pi$, which has rank $L-1$. Since $W+\mu M$ is positive definite for $\mu>0$, the pencil has exactly $L-1$ nonzero eigenvalues for every $\mu$. The objective is $v^TS_bv/(v^TS_wv+\mu\|v\|^2)=v^TS_bv/v^T(S_w+\mu I)v$, the regularized kernel Fisher discriminant. For $t=0$ the objective is $v^TCv/(v^TS_wv+\mu\|v\|^2)$, Method No.5 with a ridge.

\clearpage
\section*{S3. Results with the pooled preprocessing}
Table~\ref{tab:pooled} reports the results for GSE44076 and JAFFE identity recognition when the standardization (GSE44076) or the PCA basis (JAFFE identity) is computed on all samples before the split. The comparison with the main results table shows that the in-fold preprocessing lowers most accuracies slightly but does not change the conclusions.

\input{tables_pooled}

\end{document}

%% file: tables_results.tex
% generated by scripts/make_tables_4th.py -- do not edit by hand

\begin{table*}[t]
\centering
\caption{Test accuracy (mean (std) over folds, \%) of the seven variants and the four regularized members. Each family selects its kernel degree, subspace dimension and own parameters on the validation samples of each fold; every value is further averaged over five numerical perturbation seeds (Section~\ref{exp}). GSE44076 and JAFFE-ID use in-fold preprocessing. Best-of-7 and Best-of-4: the family with the highest validation accuracy among the seven variants and among the four regularized members, chosen in every fold. The best single family per task is in bold.}
\label{tab:main}
\renewcommand{\arraystretch}{0.9}
\setlength{\tabcolsep}{3pt}
\resizebox{\textwidth}{!}{
\begin{tabular}{lcccccc}
\toprule
Method & Indian Pines & CWRU & GSE44076 & JAFFE-EXP & JAFFE-ID & MNIST \\
\midrule
No.1 & 54.69(2.40) & 70.83(2.49) & 97.48(1.32) & 30.54(4.13) & 95.59(2.13) & 84.99(0.29) \\
No.2 & 53.44(2.20) & 71.83(1.74) & 98.45(1.17) & 28.60(6.09) & 95.05(3.92) & 87.06(0.70) \\
No.3 & 49.35(1.37) & 69.52(1.78) & 93.28(3.60) & 31.81(3.32) & 92.60(2.10) & 64.62(2.10) \\
No.4 & 52.97(2.12) & 71.89(1.72) & 98.45(1.17) & 26.91(5.48) & 94.66(3.69) & 86.86(0.87) \\
No.5 & 54.31(2.21) & 69.89(2.85) & 97.26(1.29) & 30.68(4.05) & 95.17(2.27) & 84.39(0.74) \\
No.6 & 53.44(2.43) & \textbf{72.14}(1.50) & 98.45(1.17) & 29.74(5.16) & 95.15(2.29) & 82.82(0.41) \\
No.7 & 55.07(2.56) & 70.96(3.08) & \textbf{98.64}(1.08) & 40.81(4.53) & 95.90(2.39) & 86.20(1.25) \\
\midrule
NFST ($\mu\to0^{+}$) & 53.31(3.12) & 71.39(2.84) & 98.64(1.08) & 43.44(4.25) & 97.42(2.26) & 88.51(0.82) \\
RKFD ($t\to\infty$) & \textbf{58.55}(2.23) & 71.39(3.24) & 98.64(1.08) & \textbf{43.91}(3.60) & \textbf{97.50}(2.15) & 88.17(0.86) \\
No.5$_\mu$ ($t=0$) & 57.18(1.97) & 71.04(3.15) & 98.64(1.08) & 43.47(4.03) & 97.38(2.26) & \textbf{88.83}(1.78) \\
Full ($\mu,t$) & 58.50(2.06) & 71.19(3.22) & 98.64(1.08) & 43.54(4.04) & 97.50(2.15) & 88.44(0.53) \\
\midrule
Best-of-7 & 55.36(2.07) & 70.67(2.25) & 97.53(1.37) & 37.75(6.37) & 95.81(2.00) & 86.47(1.01) \\
Best-of-4 & 58.58(2.23) & 71.44(2.87) & 98.64(1.08) & 43.47(4.18) & 97.50(2.15) & 88.41(0.61) \\
\midrule
Best-of-4 $-$ Best-of-7 & +3.22 & +0.78 & +1.11 & +5.72 & +1.69 & +1.93 \\
\quad range over seeds & [+2.96, +3.46] & [-0.83, +3.89] & [+0.58, +1.46] & [+2.94, +8.90] & [+1.07, +2.91] & [+1.17, +2.66] \\
\quad folds won & 10/10 & 7/10 & 8/10 & 9/10 & 7/10 & 4/5 \\
\bottomrule
\end{tabular}}
\end{table*}

\begin{table}[t]
\centering
\caption{The fusion ratio $t$ on the real tasks: accuracy of the member with both parameters selected (Full) minus the better of the two edges RKFD ($t\to\infty$) and No.5$_\mu$ ($t=0$), in percentage points (mean and range over seeds), and the percentage of folds in which Full selected an interior value $0<t<\infty$.}
\label{tab:t}
\renewcommand{\arraystretch}{0.9}
\setlength{\tabcolsep}{3pt}
\begin{tabular}{lcc}
\toprule
Task & Full $-$ better edge & Interior $t$ selected (\%) \\
\midrule
Indian Pines & -0.06 [-0.14, +0.09] & 60 \\
CWRU & -0.20 [-0.28, +0.00] & 8 \\
GSE44076 & +0.00 [+0.00, +0.00] & 0 \\
JAFFE-EXP & -0.47 [-1.84, +0.00] & 2 \\
JAFFE-ID & +0.00 [+0.00, +0.00] & 0 \\
MNIST & -0.39 [-1.06, -0.09] & 28 \\
\bottomrule
\end{tabular}
\end{table}

\begin{table}[t]
\centering
\caption{Median selected subspace dimension $p$ and, in parentheses, the percentage of selections at the largest admissible value (the rank of $\widetilde K$; for CWRU, 6 or more). Values over all folds and perturbation seeds.}
\label{tab:p}
\renewcommand{\arraystretch}{0.9}
\setlength{\tabcolsep}{3pt}
\resizebox{\columnwidth}{!}{
\begin{tabular}{lcccccc}
\toprule
Method & Indian Pines & CWRU & GSE44076 & JAFFE-EXP & JAFFE-ID & MNIST \\
\midrule
No.1 & 104 (54) & 2 (12) & 5 (12) & 6 (10) & 54 (50) & 559 (100) \\
No.2 & 18 (0) & 2 (0) & 1 (0) & 6 (10) & 16 (0) & 30 (0) \\
No.3 & 20 (2) & 2 (10) & 2 (2) & 8 (4) & 10 (0) & 559 (60) \\
No.4 & 18 (0) & 2 (0) & 1 (0) & 10 (20) & 16 (0) & 40 (0) \\
No.5 & 104 (68) & 3 (16) & 5 (24) & 6 (12) & 54 (50) & 559 (100) \\
No.6 & 16 (4) & 3 (10) & 1 (0) & 6 (0) & 8 (0) & 20 (0) \\
No.7 & 95 (44) & 4 (14) & 1 (0) & 7 (0) & 12 (8) & 559 (68) \\
\midrule
NFST ($\mu\to0^{+}$) & 40 (0) & 4 (22) & 1 (0) & 5 (2) & 6 (2) & 30 (0) \\
RKFD ($t\to\infty$) & 16 (0) & 3 (28) & 1 (0) & 5 (0) & 6 (0) & 50 (0) \\
No.5$_\mu$ ($t=0$) & 12 (0) & 3 (6) & 1 (0) & 5 (0) & 6 (0) & 12 (0) \\
Full ($\mu,t$) & 14 (0) & 4 (10) & 1 (0) & 5 (0) & 6 (0) & 30 (0) \\
\bottomrule
\end{tabular}}
\end{table}

\begin{table}[t]
\centering
\caption{Band (largest minus smallest mean test accuracy over five perturbation seeds, percentage points) when the feature matrices are multiplied entrywise by $1+10^{-12}\epsilon$, $\epsilon\sim\mathcal{N}(0,1)$.}
\label{tab:band}
\renewcommand{\arraystretch}{0.9}
\setlength{\tabcolsep}{3pt}
\resizebox{\columnwidth}{!}{
\begin{tabular}{lcccccc}
\toprule
Method & Indian Pines & CWRU & GSE44076 & JAFFE-EXP & JAFFE-ID & MNIST \\
\midrule
No.1 & 0.39 & 4.50 & 0.97 & 3.53 & 1.94 & 1.20 \\
No.2 & 0.00 & 0.00 & 0.00 & 0.00 & 0.00 & 0.00 \\
No.3 & 1.22 & 0.72 & 2.91 & 3.38 & 2.43 & 1.34 \\
No.4 & 0.00 & 0.00 & 0.00 & 0.00 & 0.00 & 0.00 \\
No.5 & 0.94 & 5.56 & 1.36 & 3.53 & 2.23 & 0.89 \\
No.6 & 0.26 & 0.11 & 0.00 & 0.44 & 0.97 & 0.91 \\
No.7 & 0.65 & 2.22 & 0.00 & 2.57 & 0.87 & 0.94 \\
\midrule
NFST ($\mu\to0^{+}$) & 0.37 & 1.06 & 0.00 & 2.57 & 0.10 & 1.11 \\
RKFD ($t\to\infty$) & 0.26 & 0.39 & 0.00 & 1.69 & 0.49 & 0.49 \\
No.5$_\mu$ ($t=0$) & 0.03 & 0.17 & 0.00 & 1.40 & 0.00 & 0.00 \\
Full ($\mu,t$) & 0.31 & 0.22 & 0.00 & 2.87 & 0.49 & 0.97 \\
\bottomrule
\end{tabular}}
\end{table}

\begin{table}[t]
\centering
\caption{Accuracy (\%) of different ways of choosing the family in every fold. Oracle: the family with the highest test accuracy (not implementable). Validation: the family with the highest validation accuracy. Fixed RKFD: RKFD in every fold, chosen before the experiments. Hindsight: the single family with the highest mean test accuracy on the task. The last row is the mean weighted by the number of folds (55).}
\label{tab:selection}
\renewcommand{\arraystretch}{0.9}
\setlength{\tabcolsep}{3pt}
\resizebox{\columnwidth}{!}{
\begin{tabular}{lcccccc}
\toprule
Task & \makecell{Oracle\\(11)} & \makecell{Validation\\(11)} & \makecell{Fixed\\RKFD} & \makecell{Hindsight\\(11)} & \makecell{Validation\\(7)} & \makecell{Fixed\\No.7} \\
\midrule
Indian Pines & 58.67 & 58.52 & 58.55 & 58.55 & 55.36 & 55.07 \\
CWRU & 73.50 & 70.71 & 71.39 & 72.14 & 70.67 & 70.96 \\
GSE44076 & 99.11 & 97.53 & 98.64 & 98.64 & 97.53 & 98.64 \\
JAFFE-EXP & 45.19 & 41.81 & 43.91 & 43.91 & 37.75 & 40.81 \\
JAFFE-ID & 98.72 & 96.37 & 97.50 & 97.50 & 95.81 & 95.90 \\
MNIST & 89.87 & 88.45 & 88.17 & 88.83 & 86.47 & 86.20 \\
\midrule
Weighted mean & 76.39 & 74.39 & 75.29 & 75.48 & 72.79 & 73.54 \\
\bottomrule
\end{tabular}}
\end{table}

\begin{table}[t]
\centering
\caption{Added noise $X\leftarrow X+\sigma s\,\mathcal{N}(0,1)$, with $s$ the mean feature standard deviation of the training samples. Accuracy (\%) of Method No.1, of RKFD, and of Best-of-7 and Best-of-4, and their difference. $\sigma=0$: five perturbation seeds; $\sigma>0$: three noise draws. GSE44076 and JAFFE-ID use the pooled preprocessing in this experiment.}
\label{tab:noise}
\renewcommand{\arraystretch}{0.9}
\setlength{\tabcolsep}{3pt}
\resizebox{\columnwidth}{!}{
\begin{tabular}{llccccc}
\toprule
Task & $\sigma$ & No.1 & RKFD & Best-of-7 & Best-of-4 & Difference \\
\midrule
Indian Pines & 0 & 54.69 & 58.55 & 55.36 & 58.58 & +3.22 \\
 & 0.5 & 45.99 & 52.76 & 52.27 & 53.14 & +0.87 \\
 & 1 & 39.20 & 47.98 & 47.92 & 48.27 & +0.35 \\
 & 2 & 34.95 & 40.96 & 40.98 & 41.97 & +0.98 \\
\midrule
CWRU & 0 & 70.83 & 71.39 & 70.67 & 71.44 & +0.78 \\
 & 0.5 & 65.74 & 67.06 & 66.22 & 66.91 & +0.69 \\
 & 1 & 55.22 & 57.20 & 55.93 & 56.87 & +0.94 \\
 & 2 & 44.06 & 48.35 & 44.20 & 47.74 & +3.54 \\
\midrule
GSE44076 & 0 & 98.12 & 99.18 & 98.12 & 99.18 & +1.07 \\
 & 0.5 & 97.80 & 98.90 & 97.80 & 98.90 & +1.10 \\
 & 1 & 97.35 & 98.71 & 97.35 & 98.67 & +1.33 \\
 & 2 & 96.57 & 97.93 & 96.67 & 97.93 & +1.26 \\
\midrule
JAFFE-EXP & 0 & 30.54 & 43.91 & 37.75 & 43.47 & +5.72 \\
 & 0.5 & 27.21 & 35.69 & 29.09 & 33.90 & +4.80 \\
 & 1 & 19.78 & 30.00 & 25.07 & 28.41 & +3.33 \\
 & 2 & 16.42 & 24.00 & 21.23 & 22.55 & +1.32 \\
\midrule
JAFFE-ID & 0 & 95.05 & 98.43 & 95.75 & 98.43 & +2.68 \\
 & 0.5 & 94.27 & 97.48 & 95.05 & 97.44 & +2.39 \\
 & 1 & 86.44 & 94.01 & 92.62 & 93.95 & +1.33 \\
 & 2 & 59.97 & 81.94 & 80.36 & 82.01 & +1.65 \\
\bottomrule
\end{tabular}}
\end{table}

%% file: tables_pooled.tex
% generated by scripts/make_tables_4th.py -- do not edit by hand
\begin{table}[htbp]
\centering
\caption{Test accuracy (mean (std) over folds, \%) with the pooled preprocessing, five perturbation seeds, and the band over the seeds (points).}
\label{tab:pooled}
\setlength{\tabcolsep}{3pt}
\begin{tabular}{lcccc}
\toprule
Method & GSE44076 & band & JAFFE-ID & band \\
\midrule
No.1 & 98.12(1.16) & 0.49 & 95.05(1.78) & 2.23 \\
No.2 & 98.74(1.15) & 0.00 & 94.56(4.60) & 0.00 \\
No.3 & 95.03(1.98) & 2.62 & 93.75(2.01) & 0.49 \\
No.4 & 98.74(1.15) & 0.00 & 95.15(3.36) & 0.00 \\
No.5 & 96.91(1.90) & 2.62 & 93.75(2.30) & 2.33 \\
No.6 & 98.80(1.13) & 0.10 & 95.73(2.66) & 0.58 \\
No.7 & 99.13(0.81) & 0.00 & 96.41(2.13) & 1.84 \\
\midrule
NFST ($\mu\to0^{+}$) & 99.13(0.81) & 0.00 & 98.39(2.22) & 0.39 \\
RKFD ($t\to\infty$) & 99.18(0.75) & 0.10 & 98.43(2.21) & 0.19 \\
No.5$_\mu$ ($t=0$) & 99.13(0.81) & 0.00 & 98.35(2.26) & 0.00 \\
Full ($\mu,t$) & 99.13(0.81) & 0.00 & 98.33(2.24) & 0.10 \\
\midrule
Best-of-7 & 98.12(1.16) & 0.49 & 95.75(2.02) & 2.23 \\
Best-of-4 & 99.18(0.75) & 0.10 & 98.43(2.20) & 0.19 \\
\bottomrule
\end{tabular}
\end{table}